# GEMMA-SQL: A Novel Text-to-SQL Model Based on Large Language Models

Hari Mohan Pandey[1†*], Anshul Gupta[2*], Subham Sarkar[2*], Minakshi Tomer[2*], Schneider Johannes[3], Yan Gong[1]

[1]School of Computing and Engineering, Bournemouth University, Bournemouth, United Kingdom
[2]Information Technology, Maharaja Surajmal Institute of Technology, New Delhi, India
[3]Data Science & Artificial Intelligence, University of Liechtenstein, Fürst-Franz-Josef-Strasse, Liechtenstein
hpandey@bournemouth.ac.uk and profharimohanpandey@gmail.com, anshulg1502@gmail.com, deepsub1234@gmail.com, tomer.minakshi@gmail.com, johannes.schneider@uni.li, ygong@bournemouth.ac.uk

**Abstract**: Text-to-SQL systems enable users to interact with structured databases using natural language, eliminating the need for specialized programming knowledge. In this work, we introduce GEMMA-SQL, a lightweight and efficient text-to-SQL model built upon the open-source Gemma 2B architecture. Unlike many large language models (LLMs), GEMMA-SQL is fine-tuned in a resource-efficient, iterative manner and can be deployed on low-cost hardware. Leveraging the SPIDER benchmark for training and evaluation, GEMMA-SQL combines multiple prompting strategies, including few-shot learning, to enhance SQL query generation accuracy. The instruction-tuned variant, GEMMA-SQL Instruct, achieves 66.8% Test-Suite accuracy and 63.3% Exact Set Match accuracy, outperforming several state-of-the-art baselines such as IRNet, RYANSQL, and CodeXDavinci. The proposed approach demonstrates that effective prompt design and targeted instruction tuning can significantly boost performance while maintaining high scalability and adaptability. These results position GEMMA-SQL as a practical, open-source alternative for robust and accessible text-to-SQL systems.

**Keywords**—*Domain-specific languages, Generative AI, GEMMA, Large Language Models, SPIDER, Text-to-SQL.*

## 1. Introduction

Text-to-SQL (natural language to SQL) technology plays a pivotal role in enabling non-technical users to interact seamlessly with relational databases through natural language queries [2][3]. By automatically translating natural language utterances into structured SQL queries, such systems democratize access to data analytics and empower users across various domains—including data science [4], business intelligence [2], and interactive database applications [5] to retrieve and analyse information without the need for SQL programming expertise[6]. This capability is increasingly vital as organizations continue to generate and store vast amounts of data in structured formats.

The longstanding challenge of making databases more accessible to non-specialists has inspired the development of intelligent query interfaces. Recent advances in artificial intelligence, particularly in natural language processing (NLP), have further accelerated interest in this area. Text-to-SQL models are now envisioned as core components in AI-driven data assistants capable of performing complex analytics through natural conversations.

Historically, early text-to-SQL systems relied heavily on structured database elements—such as schemas, metadata, and column indexes—to parse and map user inputs into corresponding SQL queries [7][8][9][10]. While these systems achieved some success, they required domain-specific rule engineering and struggled with generalization. The emergence of large language models (LLMs) has marked a significant turning point in this field. These models support zero-shot and few-shot prompting strategies, wherein SQL queries are generated directly from natural language using schema-aware instructions or demonstrations, without the need for extensive task-specific training [11].

---

[*] Equal contributor
[†] Corresponding author

Prompt-based methods not only enhance intuitiveness and ease of use, but also set strong baselines with minimal annotation or customization. Literature has shown that schema-sensitive prompts allow LLMs to perform structured prediction tasks such as SQL generation with high fluency and accuracy [12][13]. Moreover, few-shot prompting improves computational efficiency and mitigates the risk of overfitting associated with supervised fine-tuning. However, these models often underperform compared to their fine-tuned counterparts on benchmark datasets such as SPIDER [14] and WikiSQL [15], where generalization to unseen database schemas remains a central challenge.

In recent years, the field has experienced a paradigm shift driven by more powerful and specialized models. Notably, methods such as SQL-PaLM [13] and DIN-SQL [16], built upon advanced proprietary LLMs like PaLM-2 [17] and GPT-4 [18], have achieved state-of-the-art (SOTA) results in text-to-SQL generation. These models combine few-shot prompting with instruction tuning, schema linking, and consistency mechanisms to deliver highly accurate SQL outputs. However, several practical barriers hinder their adoption: (a) their immense parameter sizes limit deployment on low-resource or edge devices and (b) their proprietary and closed-source nature restricts transparency, reproducibility, and further adaptation to domain-specific tasks. Additionally, self-consistency techniques used in these models—where multiple outputs are generated and majority voting is applied—substantially increase inference latency, making real-time deployment difficult [19]. These issues underscore the need for efficient, open, and customizable alternatives that can match the performance of commercial-grade models while remaining lightweight and interpretable.

To address these limitations, GEMMA-SQL [20] is proposed, a novel text-to-SQL framework based on the open-source Gemma 2B model family. The Gemma architecture, designed with modularity and efficiency in mind, leverages unified technology foundations and modern ML frameworks such as Keras 3 [21], making it accessible for iterative fine-tuning even on personal computers [22]. The GEMMA-SQL model exhibits strong performance in both zero-shot and few-shot scenarios. Furthermore, GEMMA-SQL Instruct is presented, a variant fine-tuned on the SPIDER [14] dataset using schema-informed prompts and optimized with refined self-consistency mechanisms to reduce latency without sacrificing accuracy.

The key contributions of this paper are summarized as follows:

- GEMMA-SQL and GEMMA-SQL Instruct are presented, two new lightweight and open-source text-to-SQL models derived from the Gemma 2B family, offering SOTA efficiency and adaptability for low-resource environments.
- Iimprovements to the self-consistency strategy are introduced that significantly enhance SQL prediction accuracy while minimizing computational latency.
- Eextensive evaluation on the SPIDER benchmark demonstrates that the proposed models achieve competitive performance relative to leading proprietary systems, despite using a fraction of the resources.

The remainder of this paper is structured as follows: **Section 2** provides a comprehensive review of related work. **Section 3** outlines the proposed methodology, including model architecture and prompting techniques. **Section 4** presents experimental results and analysis. Finally, **Section 5** concludes the paper and discusses future directions.

## 2. Related work

This section provides a comprehensive overview of recent advancements in text-to-SQL generation, highlighting widely adopted models along with their respective advantages and limitations.

Wang et al. [23] introduced RAT-SQL, a model that emphasizes relation-aware schema encoding to enhance the understanding of database structures. By effectively capturing foreign key relationships, RAT-SQL enables the generation of more accurate SQL queries. While the model demonstrates strong performance in modeling schema relationships, its reliance on schema information alone can limit its ability to handle complex, nested queries. Building on this, Cozeman et al. extended RAT-SQL for multilingual applications, proposing mRAT-SQL+GAP to address text-to-SQL tasks in Portuguese [24]. This adaptation leverages the multilingual capabilities of BART [25] and utilizes a translated version of the Spider [14] dataset for Portuguese inputs. Their findings reveal that training on a combined English-Portuguese dataset yields better performance than training on either language individually, with the mixed dataset achieving 83% of the English-only performance (0.718 vs. 0.595). This demonstrates the effectiveness of multilingual training, particularly for low-resource languages like Portuguese.

Peters et al. contributed to the field by advancing deep contextualized word representations, though such models may fall short in fully capturing the complexity of SQL generation in cross-domain scenarios [26]. To address these limitations, RYANSQL (Recursively Yielding Annotation Network) was developed, using both textual input and database schema to generate SQL queries through a neural sketch-based slot-filling mechanism. It introduces the Statement Position Code (SPC) technique, which transforms nested SQL queries into a sequence of non-nested SELECT statements. RYANSQL excels at handling complex queries, especially on the Spider [14] benchmark dataset. However, it struggles when user queries and database schema use divergent terminologies, making schema alignment challenging.

To further enhance accuracy in complex scenarios, Zhang et al. proposed RESDSQL [27], which adopts a ranking-enhanced encoding and skeleton-aware decoding framework for text-to-SQL parsing. By decoupling schema linking and skeleton generation, RESDSQL injects the most relevant schema components into the encoder and first generates a SQL skeleton before constructing the complete query. Experimental results on Spider and its variants demonstrate the model's effectiveness and robustness, particularly in cases involving multiple schema elements and intricate logical constructs.

Early systems such as Masque/SQL [28] laid the groundwork for natural language interfaces to databases by offering basic facilities for efficient query generation. However, they struggled with complex query handling and lacked robustness in adapting to different domains. To address these challenges, the Intermediate Representation Network (IRNET) introduced a three-phase synthesis approach: beginning with schema linking, followed by the generation of an intermediate representation known as SemQ [29]. This representation acts as a bridge between natural language and SQL, improving the model's ability to generalize across domains. Inspired by lambda Dependency-based Compositional Semantics (lambda DCS), SemQL captures user intent more effectively than earlier methods [30]. However, despite these advances, it still falls short in fully bridging the gap between natural language and formal SQL syntax. Notably, it continues to struggle with modeling certain complex SQL clauses, such as INTERSECT, and remains limited in handling self-joins within the FROM clause. Addressing these limitations may require incorporating a more advanced variable-binding mechanism—potentially through lambda calculus or a scoping strategy akin to those found in Discourse Representation Structures [31].

Recent advancements in natural language processing (NLP) have made it possible to leverage neural architectures to enhance both query generation and interpretation. The introduction of attention mechanisms has enabled more precise mappings of language into logical forms, which in turn improves the interpretability of the generated queries [32]. Lin et al. further this progress by developing a grammar-based approach for Neural Text-to-SQL generation [33][34]. Traditional sequence-to-sequence models[35] often struggle with the complexities of SQL syntax and semantics, but this grammar-based method reduces over-generation while maintaining comprehensive query coverage—achieving over 98% accuracy on both the ATIS [36] and SPIDER datasets [14]. However, reliance on predefined grammar rules limits the approach's flexibility, making it less adept at handling novel or unforeseen queries. Moreover, the increased complexity introduced by this approach can make it more prone to errors, particularly when dealing with ambiguities in natural language [34].

Large Language Models (LLMs) have emerged as a powerful tool for generating SQL queries from natural language. Rajkumar demonstrated the effectiveness of zero-shot prompting, yielding reasonably good results, but acknowledged its struggles with complex queries [37]. Aiwei Liu [38] expanded on this by employing few-shot prompts, which improved accuracy but still encountered difficulties with schema linking and query complexity. To address these challenges, DIN-SQL [16] introduced a decomposition approach that breaks down the text-to-SQL task into simpler subtasks. This method not only enhances performance but also mitigates issues related to nested queries and schema ambiguities. DIN-SQL [16] reported improved execution accuracy compared to state-of-the-art methods on the Spider [14] benchmark. However, its complexity leads to higher computational costs and longer processing times, and it suffers from poor generalization when applied to out-of-domain queries. In [39]A three-stage end-to-end text-to-SQL framework is proposed that first predicts the target database using a RoBERTa-based encoder with LLM-generated rules from NLQs, and then employs critic agents to refine SQL queries, achieving improved intent prediction and generation accuracy. In [40] a novel Structure GUided text-to-SQL framework (SGU-SQL) is proposed, which incorporates syntax-based prompting and establishes structure-aware links between user queries and the database schema.

BM25 [41][42] and dense passage retrieval [43] have made significant strides in schema identification, but they still face challenges due to the diversity of natural language expressions and vocabulary mismatches. Moreover, directly integrating large database schemas into Large Language Models (LLMs) is highly inefficient, often limiting their processing capabilities and understanding [44][45]. DBCopilot addresses these issues with a flexible, two-tier copilot model that serves as a router across vast databases. The architecture consists of schema routing and SQL generation stages. It utilizes a lightweight sequence-to-sequence (Seq2Seq) neural network-based router that efficiently maps natural language questions to relevant database schemas. This model is highly adaptive, requiring minimal manual intervention to adjust to changes in the database schema. Once the schemas are identified, they are passed to an LLM to generate the final SQL queries. This approach makes DBCopilot [46] not only efficient but also scalable, surpassing retrieval-based and fine-tuning methods. Its key strengths lie in its ability to automatically learn schema knowledge, adapt to schema changes, and enhance schema-aware SQL generation through effective prompt strategies.

SQL-PaLM-2 [17] has made significant strides in SQL generation from natural language descriptions, achieving state-of-the-art (SOTA) accuracy on the Spider dataset. Its standout feature is an execution-based self-consistency prompting method, which enhances its few-shot learning capabilities [47]. This allows SQL-PaLM-2 to generate valid SQL queries with relatively few training examples, making it highly efficient. The model's contributions are notable in advancing few-shot Text-to-SQL capabilities [47], particularly with limited labeled data, through novel execution-based consistency decoding [48]. This approach helps narrow the performance gap compared to extensively trained, fine-tuned models. Additionally, SQL-PaLM-2 demonstrates robustness to the variability in natural language expressions and database schema variations. It achieves this by exploring effective learning methods, input selection strategies, and scalability. The model's findings show that more diversified training data improves generalization, relevant database content boosts performance, column selection reduces input size with minimal impact, and test-time selection optimizes results for complex SQL queries.

Table 1. A comparative analysis of Text-to-SQL models: method, dataset, key contributions, strengths, and weaknesses

| Sources | Method | Year | Dataset | Major Contributions | Strengths | Weaknesses |
|---|---|---|---|---|---|---|
| [29] | IRNet | 2019 | SPIDER[14] | Three-phase approach: schema linking, intermediate representation (SemQL), and SQL inference. Memory-augmented pointer network for schema selection. | Introduction of intermediate representation (SemQL) enhances performance in cross-domain tasks. Three-phase approach addresses intent-implementation mismatch. | Complexity of the three-phase method can increase inference time. Schema linking reliance reduces generalization to unseen schemas. |
| [49] | RYANSQL | 2020 | SPIDER[14] | Statement Position Code (SPC) for nested queries. Recursively yielding sketch-based slot filling. Input manipulation methods (JOIN Table Filtering & Supplemented Column Names). | Effective handling of nested SQL queries using SPC. High performance on SPIDER benchmark, particularly in cross-domain tasks. | Limited generalization to new schemas due to sketch-based slot filling. Frequent column selection and table classification errors, causing incorrect queries. |
| [24] | mRAT-SQL+GAP | 2021 | Portuguese Translated Spider | Adaptation for Portuguese. Multilingual approach achieves 83% of English performance. Training with English and Portuguese datasets. | Handles both English and Portuguese, demonstrating multilingual approach effectiveness. Translated datasets available for further research. | Performance drops to 83% compared to English (0.595 vs. 0.664). Struggles with Portuguese language-specific keywords. |
| [34] | Grammar SQL | 2022 | SPIDER [14] | Clause-level parallel decoding. Alignment loss for improved clause generation. Enhanced grammar-based parsing with LGESQL and RATSQL. | Efficient clause-level parallel decoding improves parsing speed. Incorporation of alignment loss improves SQL query accuracy. | Performance degrades with very large schemas. Limited flexibility for new and unseen database structures. |
| [27] | RESDSQL | 2023 | Spider[14], Spider-DK[50], Spider-Syn [51], Spider-Realistic [52] | Decouples schema linking and skeleton parsing. Ranking-enhanced encoding with schema relevance. Skeleton-aware | High robustness in handling complex scenarios with multiple schema items and logical operators. Decoupling schema and skeleton parsing improves accuracy. | Challenges in generalizing schema linking across databases. Limited scalability due to complexity of schema and skeleton parsing. |

| | | | | decoding for SQL generation. | | |
|---|---|---|---|---|---|---|
| [13] | SQL-PaLM | 2023 | SPIDER [14], Spider-Syn [51], Spider-Realistic [52], Spider-Dk [31] | Execution-based prompting. Leveraging PaLM for SQL. Robust generation of complex SQL queries. | Smaller accuracy gap for reliable SQL generation. Execution-based consistency-decoding enables strong performance with minimal training data. | Overfitting may hinder generalization. Challenges with complex SQL queries or multi-step reasoning. |
| [16] | DIN-SQL | 2023 | SPIDER[14] | Decomposed in-context learning. Adaptive prompting strategies. Self-correction for SQL queries. | Decomposition improves performance in large language models for text-to-SQL. Adaptive prompting optimizes few-shot learning. | Complexity increases computational costs and processing times. Limited generalization to new domains not covered in training data. |
| [4] | SQL-Prompt | 2023 | SPIDER[14] | Execution-based consistency decoding. MixPrompt and MixLLMs for SQL proposals. Few-shot learning with minimal labeled data. | Significant improvement in few-shot Text-to-SQL using execution-based consistency decoding. Leverages diverse prompt designs for optimization. | Reliance on multiple prompt designs and LLMs is computationally expensive and time-consuming. |
| [53] | CodeS | 2024 | SPIDER [14], BIRD [54], Spider-DK[31] ,Spider-Syn [51] Spider-Realistic [52] , Dr.Spider [55] | SQL-centric pre-training. Strategic prompt construction. Bi-directional data augmentation for domain adaptability. | Bi-directional data augmentation improves SQL generation and domain adaptability. Open-source pre-trained models for text-to-SQL tasks. | Weaknesses in handling ambiguous schema names and complex table-joins. Evaluation shows limitations with specialized datasets. |
| [46] | DBCopilot | 2024 | SPIDER [14], BIRD [54], Fiben [56], Spider-Syn [51], Spider-Realistic [52] | Schema routing and SQL generation decoupling. Seq2Seq neural network router. Reverse schema-to-question generation. | Decoupling schema routing and SQL generation improves adaptability to large-scale databases. Reverse schema-to-question generation enhances complex schema handling. | Increased system complexity due to decoupling. Synthetic data generation effectiveness may plateau, limiting improvements with unseen data. |

Traditional methods have primarily focused on fine-tuning encoder-decoder models in a supervised manner, often struggling with issues related to schema linking and domain adaptation. Recently, large language models (LLMs) such as GPT-4 and PaLM-2 have shifted the landscape, demonstrating state-of-the-art performance with minimal fine-tuning. Their key advantage lies in leveraging vast, pre-existing datasets, significantly improving generalization across diverse domains. However, these models still face a critical challenge: their reliance on closed-source architectures, which limits transparency and adaptability for specific use cases. CodeS [53] addresses several challenges, including schema linking, by employing string matching combined with neural network techniques to enhance semantic understanding. Additionally, it uses a bi-directional data augmentation strategy that generates both SQL-to-question and question-to-SQL pairs, strengthening the model's robustness to real-world queries. Despite these advancements, challenges remain in ensuring that generated queries align closely with user intent, and in safeguarding against potential data privacy concerns related to API usage.

Text-to-SQL systems continue to face several significant challenges. One of the primary obstacles is the large size of these models, which impedes efficient execution and fine-tuning, making them difficult to run on low-capacity devices and limiting their scalability. Another major issue is the difficulty in fine-tuning these models, often made nearly impossible due to their closed-source nature, which hides critical details about their architecture, training processes, and inference methods. This lack of transparency presents a substantial barrier to improving or conducting local inference, compounded by security concerns. Additionally, current methods for ensuring self-consistency in Text-to-SQL systems tend to be computationally expensive, resulting in increased latency, which undermines the feasibility of real-time applications. A further challenge is that these models often struggle with handling SQL joins and nested statements, which are essential for generating complex queries and addressing real-world use cases.

These challenges have been effectively addressed by the open-source Gemma 2B model [7], which is much smaller in size and can run locally. Built on Keras3 [21], the Gemma model allows for iterative fine-tuning over a large number of epochs without requiring excessively long training times. Its design makes it highly adaptable, even in environments with limited computational resources. Furthermore, refinements to the existing self-consistency method have significantly improved the accuracy of SQL predictions while reducing latency. This combination of size optimization, efficient fine-tuning, and enhanced query handling makes Gemma 2B a scalable, resource-efficient solution for tackling the challenges in the Text-to-SQL domain.

Table 1 presents a comparative analysis of Text-to-SQL models highlighting their methodologies, datasets used, key contributions, strengths, and limitations. During the model evaluation, several key error patterns emerged. One recurring issue was related to operation selection, where the model incorrectly inserted operators like "ascending" or "descending" when they were not appropriate. Another significant error involved counting statements, where the model often failed to correctly process queries related to counting specific data columns. A third common problem occurred when a column name in the query already existed in the schema. If the capitalization of the column name differed between the schema and the query, the model often mistakenly selected the column from the query instead of the schema. For example, when the schema used capitalized letters for column names but the query did not, the model wrongly assumed the column from the query, leading to errors in query generation. These issues highlight critical bottlenecks, including refining operation selection, improving counting logic, and ensuring accurate schema-based column identification—key areas for enhancing the model's overall accuracy and reliability in query generation.

**2.1 Rationale for Developing GEMMA-SQL Based on GEMMA-2B**

In this paper, GEMMA-SQL has been built on the Gemma 2B architecture to overcome key limitations of existing state-of-the-art text-to-SQL models. While recent advances such as SQL-PaLM [13] and DIN-SQL [16] achieve strong accuracy, they depend on large proprietary LLMs (e.g., PaLM-2, GPT-4) that are computationally expensive, closed-source, and impractical for deployment on low-resource or edge devices. Their reliance on conventional self-consistency methods also leads to high inference latency, making real-time applications difficult.

In contrast, the Gemma 2B family is open-source, modular, and highly efficient, allowing iterative fine-tuning on modest hardware while ensuring transparency and adaptability for domain-specific tasks. The 2B parameter version of GEMMA-SQL is particularly resource-friendly, capable of running on standard personal computers with far lower hardware demands than many current high-performing models. Built on Keras 3.0, the GEMMA architecture supports iterative fine-tuning across extended epochs without prolonged training times, making it practical even in low-resource environments.

Leveraging Gemma 2B, two lightweight yet competitive models, namely, GEMMA-SQL and GEMMA-SQL Instruct has been designed that enable deployment in low-resource settings, employ optimized self-consistency strategies to reduce latency without sacrificing accuracy and deliver competitive performance on the SPIDER benchmark while using only a fraction of the resources required by proprietary systems.

**3. Proposed Methodology**

In this study, a GEMMA-SQL, a text-to-SQL model is introduced adapted from Gemma [20], a lightweight, open-source model that has achieved state-of-the-art performance among models in its class. The Gemma family leverages shared technological foundations and operational frameworks, making it highly adaptable across a range of applications. GEMMA-SQL is compatible with widely used machine learning libraries, including Hugging Face Transformers, PyTorch, Keras 3.0, and JAX, offering flexibility in implementation and experimentation. The 2B parameter version of GEMMA-SQL is particularly resource-efficient, capable of running on standard personal computers with significantly lower hardware requirements compared to many current high-performing models. Built on Keras 3.0, the Gemma architecture supports iterative fine-tuning over extended epochs without the need for prolonged training times, an approach that remains viable even on low-resource devices. Initial evaluations show that GEMMA-SQL performs effectively with minimal or no in-context examples and further improves when using few-shot prompting. Moreover, fine-tuning the model on the SPIDER dataset [14], followed by the application of prompting strategies such as zero-shot and few-shot learning, has led to notable enhancements in both accuracy and overall performance.

Large Language Models (LLMs) have achieved performance on par with state-of-the-art systems across various natural language processing (NLP) tasks, largely due to comprehensive pre-training on diverse and extensive corpora. Despite their general capabilities, domain-specific tasks such as Text-to-SQL require task-oriented fine-tuning to enhance prompt interpretation and improve the accuracy of structured query generation. In this study, the Gemma model was fine-tuned on the SPIDER [14] training dataset, which includes paired natural language questions and database schemas as inputs, with the objective of producing corresponding SQL queries. The fine-tuning process yielded significant improvements in Text-to-SQL performance, underscoring the necessity of domain adaptation. This section details the methodology employed for model fine-tuning and subsequent evaluation.

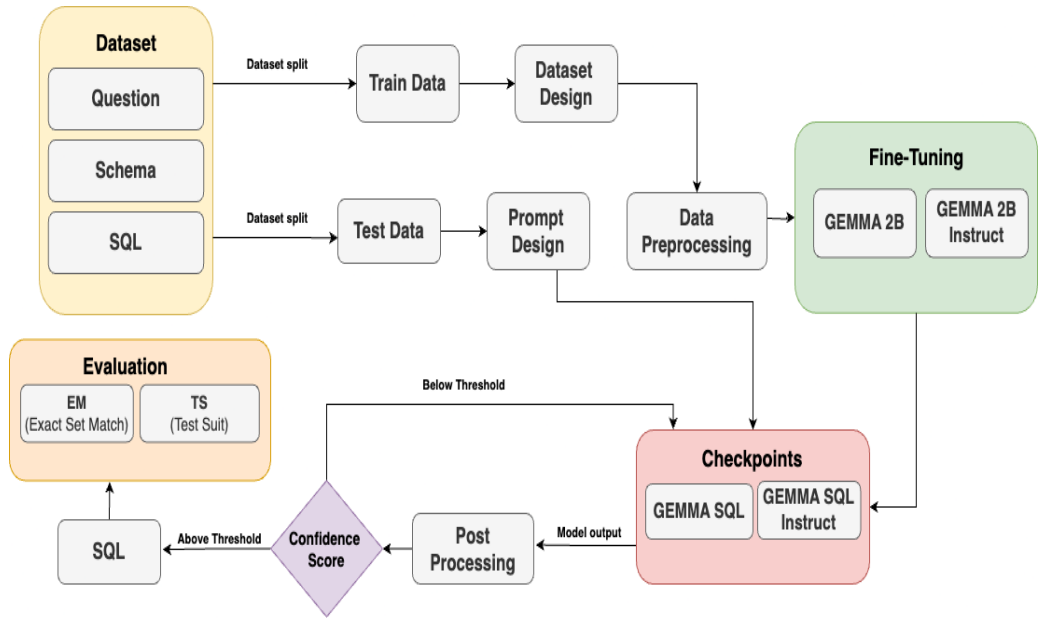

**Figure 1**. The diagram illustrates the fine-tuning pipeline for SQL query generation, featuring an iterative refinement process guided by confidence score thresholds.

Figure 1 presents an end-to-end, phase-wise framework for fine-tuning and evaluating Large Language Models (LLMs), such as Gemma, in the context of the Text-to-SQL generation task. The pipeline is structured into several interconnected phases, each designed to ensure optimal model training, prompt formulation, inference, post-processing, and evaluation.

*Dataset Preparation Phase:* The process begins with a curated dataset consisting of natural language questions, associated database schemas, and ground-truth SQL queries. This dataset is partitioned into two subsets: a training set and a test set. The training set is used to construct structured input representations, aligning the data format with the requirements of LLM-based fine-tuning. This includes organizing the schema information and question-SQL pairs into a form that maintains relational context.

*Fine-Tuning Phase*: Once the training data is structured, it is passed through a preprocessing stage to generate tokenized and context-aware input for model training. The preprocessed data is then used to fine-tune two variants of the LLM—GEMMA 2B and GEMMA 2B Instruct. These variants differ in their architectural instructions and supervision signals. Fine-tuning enables the models to better capture the nuances of SQL query generation from natural language prompts, particularly in aligning question semantics with underlying database schemas.

*Prompt Engineering and Inference Phase*: Simultaneously, the test data is used for prompt design, where task-specific prompt templates are crafted to guide the model during inference. These prompts emulate realistic user inputs and are essential for prompting the model to generate contextually accurate SQL queries [57]. The prompt-structured test inputs are then passed to the fine-tuned model checkpoints—GEMMA SQL and GEMMA SQL Instruct—which represent the saved states of the fine-tuned models. These checkpoints generate SQL outputs corresponding to each test query.

*Post-Processing and Confidence Scoring Phase*: The generated SQL queries undergo a post-processing stage where outputs are cleaned, normalized, or revised to ensure syntactic and semantic correctness. Each query is then assigned a confidence score, which reflects the model's certainty in the generated response. This score is calculated based on internal model heuristics or external validation criteria. If the confidence score exceeds a predefined threshold, the output is considered for final evaluation. If it falls below the threshold, the query is routed through an iterative refinement loop, possibly involving additional post-processing or re-inference via model checkpoints.

*Evaluation Phase*: Final outputs that pass the confidence score filter are evaluated using two standard metrics: Exact Match (EM), which checks whether the predicted SQL query matches the ground-truth query exactly, and Test Suite (TS), which validates the functional equivalence of the predicted SQL by executing it on the schema and comparing the results. These metrics collectively assess both syntactic accuracy and functional correctness, offering a robust evaluation of model performance.

Overall, the framework in Figure 1 exemplifies a modular, data-driven approach to optimizing LLMs for the Text-to-SQL task. It combines principled dataset handling, targeted model fine-tuning, prompt engineering, iterative feedback through confidence scoring, and comprehensive evaluation. This design ensures not only the adaptability of the LLM to structured query generation but also continuous performance enhancement through informed feedback loops. The subsequent subsections elaborate on each component of this architecture, while Section 4 provides detailed insights into the SPIDER [14] dataset used for model training and evaluation.

### 3.1 Problem Setup

The two fine-tuned model variants—GEMMA-SQL and GEMMA-SQL-Instruct—transform user-provided database schemas ($S$) and natural language queries ($Q$) into executable SQL statements. The underlying database is formally represented as $D \coloneqq (S, K_p, K_f)$, where $S$ denotes the schema, $K_p$ the set of primary keys, and $K_f$ the set of foreign keys. The schema $S$ comprises a collection of tables $T = \{T_1, T_2, \ldots, T_n\}$, where each table $T_t$ is characterized by a table name $N_t$ a set of column names $\{C_j\}$, and their corresponding data types $\{t_j\}$ For every table, the schema must explicitly define the structure, including column–data type pairs and constraints. Additionally, essential metadata such as the database name, primary keys (which uniquely identify rows), and foreign key relationships (which capture inter-table dependencies) are included to fully describe the relational context. This comprehensive representation ensures that the model has sufficient contextual information to accurately generate SQL queries aligned with the schema's structural and semantic constraints.

### 3.2 Base LLM: GEMMA

Gemma leverages a transformer-based neural architecture designed to effectively capture intricate relationships and long-range dependencies within textual data. The model pipeline begins with tokenization of the input text, which is subsequently embedded into high-dimensional numerical vectors. These embeddings are then processed through multiple transformer layers, each comprising multi-head self-attention mechanisms and position-wise feed-forward neural networks. The attention mechanism enables the model to attend to diverse parts of the input simultaneously, thereby facilitating the learning of complex linguistic patterns and semantic associations. A key innovation in Gemma is the integration of rotary positional embeddings (RoPE), a method that encodes positional information more efficiently than traditional approaches. RoPE enhances the model's contextual understanding across varying input lengths while simultaneously reducing the overall model size, leading to improved memory and computational efficiency. This design choice provides greater flexibility in handling sequences of different lengths without compromising coherence or relevance. Furthermore, multi-query attention is employed to enhance performance at various model scales by optimizing the attention computation process, thereby reducing latency and improving throughput. The final output representations, generated at the terminal layers of the transformer stack, can be decoded into human-readable text or adapted for downstream language tasks such as question answering, dialogue generation, and SQL query synthesis. To ensure robustness and domain adaptability, Gemma undergoes fine-tuning on domain-specific datasets and is rigorously evaluated across multiple benchmarks. This process not only boosts performance but also strengthens generalization and safety, enabling the model to be effectively deployed in diverse application scenarios.

### *3.3 Dataset Preparation*

The proposed models utilize an iterative data preparation approach in which each row of the dataset is processed to extract values from the 'question', 'SQL', and 'schema' columns. These values are then formatted into a unified string structure, as defined in Equation (1), to support a self-supervised fine-tuning framework.

$$\text{String Template: \{Instruction: question, Schema: schema, Response: SQL\}} \qquad (1)$$

This structured template concatenates the natural language instruction, the associated database schema, and the corresponding SQL query, using clearly defined labels. The use of such a template standardizes the input-output representation for the model, enhancing its ability to learn consistent mappings between questions and SQL responses under varying schema conditions.

To train the model effectively, a total of 8,000 data points were generated using the format described in Equation (1). An illustrative example of a single training instance constructed using this method is shown in Figure 2, providing a concrete visualization of how the template is applied in practice.

```
# Instruction:
'what is the biggest city in wyoming' ?

# Schema:
{'database': 'geo',
 'metadata': [{'name': 'city',
    'columns': ['city_name', 'population', 'state_name']}]}

# Response:
'SELECT city_name FROM city WHERE population  =  ( SELECT MAX ( population ) FROM city WHERE state_name  =  "wyoming" ) AND state_name  =  "wyoming";'
```

**Figure 2**. An example of a single data point used for fine-tuning the model through a self-supervised learning approach.

Figure 2 illustrates an example of a single data point constructed using the string template described in Equation (1). In this instance, the instruction component contains the natural language query: "*What is the biggest city in Wyoming?*". The Schema section defines the structure of the relevant database, including the database name ("geo") and its metadata. The metadata describes a table named "city", which includes three columns: *city_name*, *population*, and *state_name*.

The response portion contains the corresponding SQL query that answers the instruction. Specifically, the SQL command selects the *city_name* from the city table where the population is the maximum within the state of Wyoming. The query employs a subquery to compute the maximum population and uses it to filter the desired city.

This example demonstrates the structured alignment of the natural language input, database schema, and SQL output, showcasing how the model is trained to learn mappings from text to executable SQL queries under schema constraints.

### *3.4 Preprocessing*

Pre-processing begins by removing extraneous elements such as unnecessary newlines, user tags (e.g., @user), and HTML or Markdown syntax to ensure a clean input for further processing. The dataset is initially loaded into a pandas Data Frame and subsequently converted into a string format suitable for input into the GEMMA-SQL model during fine-tuning. This input string comprises two key components: the *Instruction* and the *Schema*. GEMMA-SQL uses these components to generate a corresponding SQL query.

```
#Instruction:
"What is the paper about convolution from brian curless ?"
```

```
#Schema:
{'database': 'scholar', 'metadata': [{'name': 'author', 'columns': ['authorid', 'authorname']},
{'name': 'keyphrase', 'columns': ['keyphraseid', 'keyphrasename']},
{'name': 'paper', 'columns': ['paperid']}, {'name': 'paperkeyphrase', 'columns': ['paperid', 'keyphraseid']},
{'name': 'writes', 'columns': ['paperid', 'authorid']}]}
```

```
#Response:
SELECT DISTINCT t1.authorid, t3.paperid
FROM paperkeyphrase AS t2
JOIN keyphrase AS t5 ON t2.keyphraseid = t5.keyphraseid
JOIN paper AS t3 ON t3.paperid = t2.paperid
JOIN writes AS t4 ON t4.paperid = t3.paperid
JOIN author AS t1 ON t4.authorid = t1.authorid
WHERE t1.authorname = "brian curless" AND t5.keyphrasename = "convolution";
```

**Figure 3.** Model-generated SQL output obtained through schema-based preprocessing for accurate query extraction

Figure 3 illustrates this process with an example, where the natural language query "*What is the paper about convolution from Brian Curless?*" is transformed into a structured SQL query. The underlying schema represents a scholarly database with entities such as *author*, *keyphrase*, *paper*, *paperkeyphrase*, and *writes*, each defined by specific attributes (e.g., *author* includes *authorid* and *authorname*, while *keyphrase* includes *keyphraseid* and *keyphrasename*). The model processes the instruction and generates a SQL query that joins the relevant tables to extract the desired information. Specifically, it retrieves the *authorid* and *paperid* for papers authored by "Brian Curless" and tagged with the keyphrase "convolution." This is achieved through a series of table joins—connecting *paperkeyphrase* with *keyphrase*, *paper*, *writes*, and finally *author*—along with appropriate filtering conditions.

*3.5 Fine Tuning*

Large-scale pre-training and instruction fine-tuning have established that LLMs exhibit outstanding performance across a wide array of tasks. However, achieving optimal results in specialized domains often requires task-specific adaptation—one such example being the conversion from natural language text to structured SQL queries. This paper investigates the fine-tuning of LLMs for Text-to-SQL tasks, utilizing the GEMMA models on the SPIDER development dataset, a benchmark specifically designed for evaluating Text-to-SQL models.

Fine-tuning LLMs, however, is a resource-intensive process, requiring significant computational power and memory. To address these challenges, we leverage low-rank adaptation [58] (LoRA), a method that improves the efficiency of model adaptation by approximating weight updates through low-rank decomposition. LoRA reduces the number of trainable parameters, allowing the model to be fine-tuned with fewer resources while maintaining its performance. This approach not only keeps inference times rapid but also allows for flexible adaptation to a wide range of tasks, all through a single pre-trained model equipped with multiple LoRA modules. Moreover, LoRA plays a crucial role in enabling the adaptation of models on hardware with more modest capabilities by reducing the memory overhead typically required for storing optimizer states. By minimizing the computational burden without sacrificing performance, LoRA enhances the scalability of LLMs, making them more accessible for specialized tasks such as Text-to-SQL conversion.

Experiments show that LoRA has proven to be highly effective in parameterizing and deploying LLMs for specific domains, such as the complex task of converting natural language into SQL queries. The results demonstrate that LoRA's efficient fine-tuning process facilitates the use of large pre-trained models for domain-specific applications without requiring extensive computational resources, thus making LLMs more practical for real-world deployment in specialized tasks.

### 3.6 Prompt Design using Few-Shot Learning

The model leverages the principles of few-shot learning [55], wherein prompts are constructed by prefixing the target natural language question with a set of illustrative examples consisting of an instruction, schema, and corresponding response. This strategy significantly enhances model performance during evaluation by providing contextual cues that guide the generation process. Moreover, the use of few-shot prompting offers a robust defence against overfitting. Since the model is not explicitly fine-tuned on a large number of labeled training examples, the risk of overfitting to the training set is substantially reduced, leading to more realistic and generalizable performance estimates.

Additionally, this method fosters improved generalization capabilities. By exposing the model to a diverse range of examples within the prompts, the model learns to adapt its internal representations and apply its acquired knowledge to novel, unseen queries. This ability to generalize from limited, representative examples is particularly valuable in real-world applications where labeled training data may be scarce or heterogeneous. As a result, the few-shot learning approach not only boosts performance but also ensures the model remains adaptable and reliable across varied Text-to-SQL scenarios.

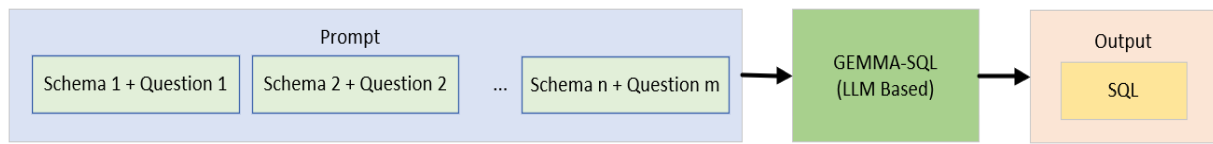

**Figure 4**. Schema-based prompting process for SQL Generation using a LLM.

```
Template = """
You are SQL developer, an developer with exceptional capabilities in understanding and generating SQL queries.
SQL Developer will be given a database schema and a natural language question in English by database administrator,
and will have to generate a SQL query to answer the question, following these rules:

SQL Developer will always provide a SQL query as an answer, even if it's uncertain or doesn't fully understand the question.
When providing the answer, SQL Developer will ONLY provide the SQL query, no more and no less.
SQL Developer is guaranteed to produce a correct SQL query and is always certain of the query's correctness.
The SQL query provided by SQL Developer must retrieve the requested information based on the given schema and question.
The SQL query must adhere to standard SQL syntax and should be compatible with commonly used
relational database management systems (e.g., MySQL, PostgreSQL, SQLite).

# Instruction:
'what year had the most NIPS papers' ?

# Schema:
{'database': 'scholar',
 'metadata': [{'name': 'venue', 'columns': ['venueid', 'venuename']},
  {'name': 'paper', 'columns': ['paperid', 'venueid', 'year']}]}

# Response:
SELECT DISTINCT COUNT ( t1.paperid ) , t1.year FROM venue AS t2 JOIN paper AS t1 ON
t2.venueid = t1.venueid WHERE t2.venuename = "NIPS" GROUP BY t1.year ORDER BY COUNT ( t1.paperid ) DESC;

# Instruction:
'What has Richard Ladner published at chi ?'

# Schema:
{'database': 'scholar',
 'metadata': [{'name': 'venue', 'columns': ['venueid', 'venuename']},
  {'name': 'author', 'columns': ['authorid', 'authorname']},
  {'name': 'paper', 'columns': ['paperid', 'venueid']},
  {'name': 'writes', 'columns': ['paperid', 'authorid']}]}

# Response
'SELECT DISTINCT t3.paperid FROM venue AS t4 JOIN paper AS t3 ON t4.venueid = t3.venueid JOIN writes AS t2 ON
t2.paperid = t3.paperid JOIN author AS t1 ON t2.authorid = t1.authorid WHERE t1.authorname = "Richard Ladner" AND
t4.venuename = "chi";'
```

**Figure 5**. Prompting strategy for generating SQL query from instruction and schema.

Figure 4 illustrates the high-level architecture of the GEMMA-SQL pipeline for SQL generation using LLMs. The input prompt is constructed using a series of example pairs, each consisting of a database schema and a corresponding natural language question (e.g., Schema 1 + Question 1, Schema 2 + Question 2, …, Schema n + Question m). These exemplars serve as context for the LLM to learn the mapping from natural language to SQL. The prompt is fed into the GEMMA-SQL model, which is based on a lightweight, fine-tuned LLM. The model processes the structured prompt and generates an appropriate SQL query as output, aligned with the schema and intent of the final question. This approach leverages in-context learning through few-shot prompting to improve the model's ability to generalize to unseen questions and schemas.

Figure 5 illustrates the process in which a combination of an instruction or question and its corresponding database schema is provided as input, and the model generates the appropriate SQL query as output.

As discussed, GEMMA-SQL employs few-shot prompting strategies to bridge natural language (NL) instructions and structured SQL queries. These strategies are carefully designed to improve both the accuracy of SQL generation and the model's adaptability to unseen schemas.

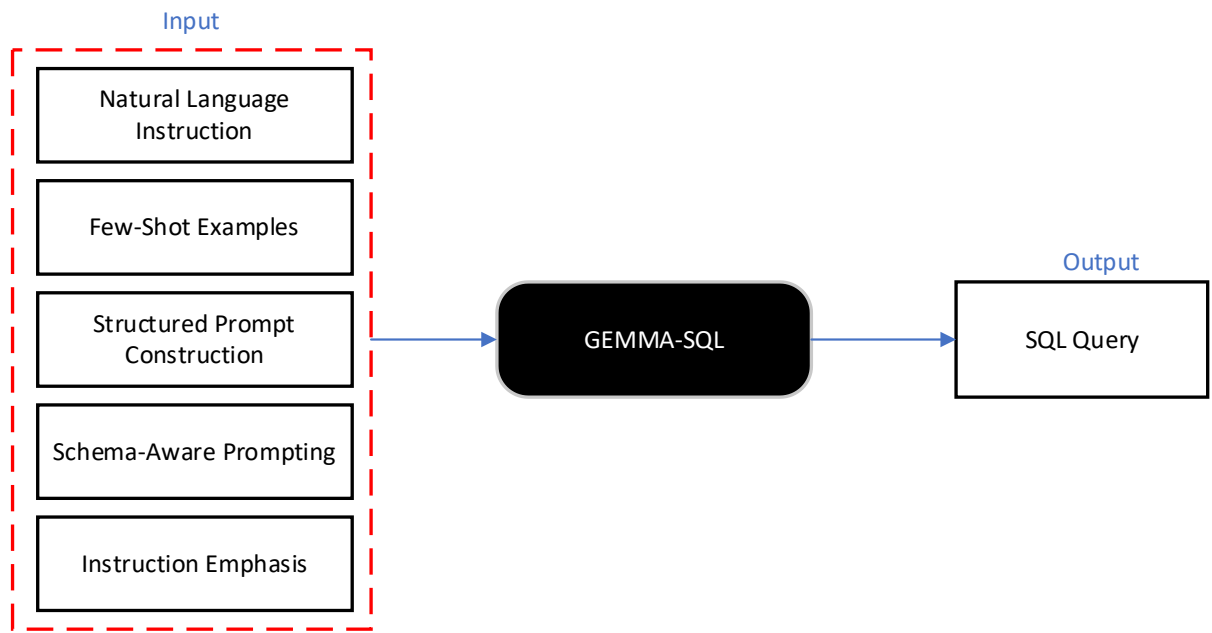

Figure 6. An overall pipeline of GEMMA-SQL.

The Figure 6 represents the overall pipeline of GEMMA-SQL, highlighting how natural language queries are systematically transformed into valid SQL statements. The process begins with a natural language instruction, such as "*List all employees earning more than 50k.*" On its own, a plain instruction may not be sufficient for accurate SQL generation, which is why few-shot examples are included in the prompt. For instance, an example might show that the instruction "*Find the names of employees earning above 50k*" with schema *Employees(id, name, salary)* maps to the SQL query *SELECT name FROM Employees WHERE salary > 50000;*. These examples help GEMMA-SQL understand the mapping between natural language and SQL. To further guide the model, structured prompt construction is used, where every example follows the same sequence of *instruction → schema → SQL*, ensuring consistency and reducing ambiguity.

Another important input is schema-aware prompting, which explicitly provides the schema of the database, including table names, columns, and relationships. For example, if the schema consists of *author(authorid, authorname)*, *paper(paperid, venueid, year)*, and *venue(venueid, venuename)*, GEMMA-SQL can use this knowledge to generate a query as shown in Figure 7.

```sql
SELECT t1.year
FROM venue AS t2
JOIN paper AS t1 ON t1.venueid = t2.venueid
WHERE t2.venuename = 'NIPS'
GROUP BY t1.year
ORDER BY COUNT(t1.paperid) DESC;
```

**Figure 7**. Demonstration of schema-aware query generation using GEMMA-SQL.

In response to the instruction "*Find the year with the most NIPS papers.*" Similarly, instruction emphasis helps refine the intent of the query by stressing constraints. For instance, the instruction "*Get the number of papers Richard Ladner published at CHI*" leads GEMMA-SQL to generate SQL query as depicted in Figure 8.

```sql
SELECT COUNT(*)
FROM author a
JOIN writes w ON a.authorid = w.authorid
JOIN paper p ON w.paperid = p.paperid
JOIN venue v ON p.venueid = v.venueid
WHERE a.authorname = 'Richard Ladner'
AND v.venuename = 'CHI';
```

**Figure 8**. Instruction emphasis query generation using GEMMA-SQL.

At the core of this pipeline, GEMMA-SQL leverages in-context learning exploitation, meaning it does not require retraining on new data but instead adapts using the examples and schema provided in the prompt. The final output is a syntactically correct and semantically meaningful SQL query that can be executed directly on the database. This structured approach makes GEMMA-SQL flexible, reliable, and highly effective in translating diverse natural language instructions into database queries across different schemas.

Algorithm 1 presents pseudocode that illustrates the GEMMA-SQL workflow and highlights the step-by-step application of few-shot prompting.

| Algorithm-1: GEMMA-SQL Few-Shot Prompting Workflow |
|---|
| **Input**: |
| NL_question      // Target natural language question |
| Example_set      // Set of illustrative examples (instruction/question, schema, SQL query) |
| **Output**: |
| SQL_query      // Generated SQL query aligned with target instruction and schema |
| 1.     Begin |
| 2.       Input Preparation |
| 3.         Collect NL_question |
| 4.         Select a subset of illustrative examples from Example_set |
| 5.         For each example in Example_set: |
| 6.           Ensure example includes: instruction/question, database schema and SQL query |
| 7.         End For |

8.     Prompt Construction
9.       Initialize Prompt ← empty string
10.      For each example in selected examples:
11.         Append example (instruction + schema + SQL query) to Prompt
12.      End For
13.      Append NL_question to Prompt as the target query instruction
14.    Model Processing
15.      Feed Prompt into GEMMA-SQL
16.      GEMMA-SQL uses in-context learning to interpret mappings from NL_question to SQL
17.      Receive preliminary SQL_query from GEMMA-SQL
18.    SQL generation and output validation
19.      If SQL_query is syntactically correct AND aligns with schema:
20.         Proceed
21.      Else
22.         Apply error correction or regeneration mechanisms
23.      End If
24.    Results
25.      Return SQL_query as the final output
26. End

***Step-by-step explanation of Algorithm-1***: GEMMA-SQL few-shot prompting workflow begins with the preparation of inputs. The process starts by collecting the target natural language question (NL_question) that the user wants to translate into SQL. Alongside this, a subset of illustrative examples is selected from the available Example_set. Each example must be complete, containing three essential components: the instruction or question in natural language, the relevant database schema with its tables and fields, and the corresponding SQL query that correctly addresses the instruction. This ensures that the examples provided to the model are structured and contextually accurate.

Once the inputs are ready, the workflow moves into the prompt construction phase. An initially empty prompt is created, and for each of the selected examples, the instruction, schema, and SQL query are appended in order. This builds a structured prompt that demonstrates to the model how natural language maps to SQL. After all examples are included, the target NL_question is appended at the end of the prompt, clearly marking it as the query to be translated.

In the model processing stage, the constructed prompt is fed into GEMMA-SQL. The model leverages in-context learning, interpreting the relationships shown in the examples to map the new NL_question into a corresponding SQL query. At this stage, a preliminary SQL_query is generated by GEMMA-SQL.

The next step is SQL generation and validation. Here, the workflow checks whether the produced SQL_query is both syntactically correct and aligned with the given schema. If the query satisfies these conditions, it is accepted as correct. Otherwise, error-handling mechanisms are triggered, which may involve refining the prompt, adding additional examples, or regenerating the query until it meets the validation criteria.

Finally, in the results stage, the validated SQL_query is returned as the final output. This ensures that the user receives a reliable query that can be directly executed on the database. Through this structured workflow, GEMMA-SQL efficiently transforms natural language instructions into precise SQL statements without the need for retraining, relying instead on few-shot prompting and in-context learning.

***Analysing computation cost of Algorithm-1***: The computational cost of the GEMMA-SQL few-shot prompting workflow can be analysed in terms of asymptotic notation by considering each stage of the algorithm. Let $E$ denote the size of the full example set, $k$ the number of selected examples, $L_e$ the average token length of one example and $L_q$ the length of the target natural language question. The total prompt length can be expressed as $L = k \cdot L_e + L_q$.

The first step, example selection, can vary in cost depending on the approach. A simple random sampling or naive scan incurs a cost of $O(E)$, whereas similarity-based retrieval with precomputed embeddings also scales as $O(E)$. Prompt construction, which involves concatenating $k$ examples and the target question, scales linearly with the total token length, giving a cost of $O(L)$.

The dominant cost arises during model inference. GEMMA-SQL uses a transformer-based architecture, and standard self-attention has a per-layer time complexity of $O(L^2 \cdot d)$, where $d$ is the hidden dimension. Across $H$ layers, this becomes $O(H \cdot d \cdot L^2)$. For practical purposes, with model size fixed, this is often simplified to $O(L^2)$. SQL validation and syntax checking scale with the length of the generated SQL query and schema size, which can be approximated as $O(L_{sql})$. If error correction or regeneration mechanisms are applied, and up to $t$ attempts are allowed, the total inference and validation cost is multiplied by $t$, resulting in $O(t \cdot (L^2 + L_{sql}))$.

Overall, the total time complexity of the algorithm, combining example selection, prompt construction, and multiple inference attempts, can be expressed as $O(E + L + t \cdot L^2 + t \cdot L_{sql})$. With fixed model size and typically $L$ is much greater than $L_{sql}$, this simplifies to $O(E + t \cdot L^2)$. If example selection is constant-time or uses pre-indexed retrieval, the cost reduces further to $O(t \cdot L^2)$.

In terms of space complexity, storing the prompt requires $O(L)$ space, while maintaining model activations during inference requires $O(H \cdot L \cdot d)$. Model parameters contribute $O(P)$ space, which is independent of the prompt length. In practical scenarios, the inference step dominates runtime, and optimisations such as reducing the number of examples, using efficient attention mechanisms, caching embeddings, and limiting regeneration iterations can significantly reduce computational overhead.

The GEMMA-SQL workflow's asymptotic computational cost is dominated by the transformer inference stage, typically $O(t \cdot L^2)$ for standard attention, with additional $O(E)$ cost for example selection if scanning the full dataset, while space complexity is $O(P + H \cdot L \cdot d)$.

***Practical example (to ground the big-O):*** Assume the following parameters for the GEMMA-SQL workflow: the total example set size is $E = 1000$, the number of selected examples is $k = 5$, the average token length of each example is $L_e = 300$ tokens, and the target question length is $L_q = 30$ tokens. This gives a total prompt length of $L = k \cdot L_e + L_q = 5 \cdot 300 + 30 = 1530$ tokens.

The cost of prompt construction, which involves concatenating the selected examples and the question, is $O(1530)$, which is negligible compared to the model inference cost. One forward pass through the transformer with standard self-attention has a complexity of $O(L^2) = O(1530^2) \approx 2.3$ million token-pair operations per layer, ignoring the number of layers and hidden dimensions for simplicity. If up to $t = 3$ regeneration or error-correction attempts are performed, the total inference cost scales linearly with the number of attempts, giving approximately $3 \times 2.3$ millions token-pair operations.

From this analysis, it is clear that model inference dominates the overall computational cost. Consequently, reducing the number of selected examples $k$ or using efficient attention mechanisms can drastically reduce runtime, while prompt construction and SQL validation contribute relatively little to the total cost.

Two examples are presented to demonstrate the working of Algorithm 1. Example 1 illustrates a case where the generated SQL query is correct, and no error correction is required. In contrast, Example 2 demonstrates a scenario that involves error correction.

**Example 1**: The SQL query generated is correct, so no error correction is needed.

| NL_question (Target Query) | : | *"List all employees earning more than 50k."* |
|---|---|---|
| Example_set | : | Contains pairs of natural language instructions, database schemas, and SQL queries. |

Step-1: Input Preparation

The workflow begins by collecting the target NL_question: *"List all employees earning more than 50k."* From the Example_set, a few relevant examples are select that demonstrate how natural language queries map to SQL. For instance:

| Example 1: | ➢ Instruction: *"Find the names of employees working in the Sales department."*<br>➢ Schema: Employees(id, name, department, salary)<br>➢ SQL: SELECT name FROM Employees WHERE department = 'Sales'; |
|---|---|
| Example 2: | ➢ Instruction: *"Retrieve employees with a salary below 40k."*<br>➢ Schema: Employees(id, name, department, salary)<br>➢ SQL: SELECT name FROM Employees WHERE salary < 40000; |

These examples form the foundation for guiding the GEMMA-SQL model.

Step-2: Prompt Construction

Next, structured prompt will be designed. It starts with an empty string and append the selected examples in the consistent format: Instruction → Schema → SQL query. The constructed prompt is presented in Figure 9.

```
Instruction: Find the names of employees working in the Sales department.
Schema: Employees(id, name, department, salary)
SQL: SELECT name FROM Employees WHERE department = 'Sales';

Instruction: Retrieve employees with a salary below 40k.
Schema: Employees(id, name, department, salary)
SQL: SELECT name FROM Employees WHERE salary < 40000;

Instruction: List all employees earning more than 50k.
Schema: Employees(id, name, department, salary)
SQL:
```

**Figure 9**. Constructed prompt.

Notice how the target NL_question is added at the end, with SQL left blank for GEMMA-SQL to generate.

Step-3: Model Processing

The constructed prompt is then fed into GEMMA-SQL. Using in-context learning, the model observes the few-shot examples and learns the mapping from natural language to SQL in this schema. For the target question, GEMMA-SQL generates the preliminary SQL query as shown in Figure 10.

```
SELECT name FROM Employees WHERE salary > 50000;
```

**Figure 10**. GEMMA-SQL generated preliminary SQL query.

Step-4: SQL Generation & Validation

The generated SQL is now validated. First, a syntactic check confirms that the query follows SQL grammar rules. Next, a schema alignment check verifies that all fields (name, salary) exist in the schema *Employees(id, name, department, salary)*. Since the query is valid, no error correction is required.

Step-5: Result

The final output of the Algorithm-1 is the validated SQL query as shown in Figure 11.

```
SELECT name FROM Employees WHERE salary > 50000;
```

Figure 11. Validated SQL query.

This query can be directly executed on the database to retrieve the expected results.

In this way, Algorithm-1 combines input preparation, structured prompting, model inference, and validation to reliably translate natural language into SQL queries using few-shot prompting with GEMMA-SQL.

**Example 2**: Let's extend the step-by-step walkthrough of Algorithm-1 with an error case to show how iterative error mitigation works.

| NL_question (Target Query) | : | "Find the total salary of all employees in the Sales department." |
| Schema | : | Employees(id, name, department, salary) |
| Example_set | : | Same as before (contains NL → Schema → SQL mappings). |

Step-1 Input Preparation

As before, the NL_question has been collected and select illustrative examples from the Example_set.

| Example 1: | • Instruction: *"Find the names of employees working in the Sales department."*<br>• Schema: Employees(id, name, department, salary)<br>• SQL: SELECT name FROM Employees WHERE department = 'Sales |
| Example 2: | • Instruction: *"Retrieve employees with a salary below 40k."*<br>• Schema: Employees(id, name, department, salary)<br>• SQL: SELECT name FROM Employees WHERE salary < 40000; |

Step-2: Prompt construction

The structured prompt is built as before, appending examples in order (Instruction → Schema → SQL). At the end, the target NL_question is appended as shown in Figure 12.

```
Instruction: Find the total salary of all employees in the Sales department.
Schema: Employees(id, name, department, salary)
SQL:
```

Figure 12. Generated prompt with target NL_question.

Step-3: Model Processing (Initial Attempt)

GEMMA-SQL processes the prompt and generates a preliminary query as presented in Figure 13.

```
SELECT SUM(salary) FROM Employee WHERE dept = 'Sales';
```

Figure 13. Preliminary SQL query generated by GEMMA-SQL.

Step-4: SQL Generation & Validation

At this stage, the generated SQL query undergoes validation to ensure both syntactic correctness and conformity with the underlying database schema. Although the query is syntactically correct, an error is identified during the schema alignment check. Specifically, two issues are observed: (a) the table name Employee does not exist in the schema (the correct table name is Employees); and (b) the field dept is invalid (the correct field name is department). Since the generated SQL query fails to align with the schema, corrective modification is required before execution.

Step-5: Error Mitigation via Iterative Prompting

To address the error, the workflow refines the prompt by either (a) incorporating explicit clarifications (e.g., 'Use the exact field names from the schema: department, salary'), or (b) introducing an additional illustrative example that emphasizes correct schema usage. Therefore, the generated prompt snippet is updated, as illustrated in Figure 14, while Figure 15 presents the corrected SQL query obtained after reprocessing through GEMMA-SQL.

```
Instruction: Find the average salary of employees in the HR department.
Schema: Employees(id, name, department, salary)
SQL: SELECT AVG(salary) FROM Employees WHERE department = 'HR';

Instruction: Find the total salary of all employees in the Sales department.
Schema: Employees(id, name, department, salary)
SQL:
```

Figure 14. Updated prompt snippet.

```
SELECT SUM(salary) FROM Employees WHERE department = 'Sales';
```

Figure 15. Correct SQL query after reprocessing by GEMMA-SQL.

Step-6: Result

The validated query is now syntactically correct and aligned with the schema. It is returned as the final output as shown in Figure 15.

Example 2 demonstrates how Algorithm 1 incorporates an iterative error-handling mechanism. When GEMMA-SQL generates a query that is syntactically valid but inconsistent with the database schema, the workflow initiates a refinement process. This process involves either augmenting the prompt with explicit clarifications (e.g., enforcing the use of exact field names) or introducing additional illustrative examples that emphasize correct schema usage. The refinement cycle is repeated until the regenerated output achieves both syntactic validity and full schema conformity, thereby ensuring the reliability of the final SQL query.

GEMMA-SQL employs few-shot prompting strategies to bridge natural language instructions and structured SQL queries. These strategies are carefully designed to improve both the accuracy of SQL generation and the model's adaptability to unseen schemas. Table 2 summarizes the principles, strategies, benefits, and corresponding SQL examples for each GEMMA-SQL prompting approach.

**Table 2.** An overview of GEMMA-SQL prompting approaches, detailing their principles, implementation strategies, benefits and illustrative SQL query examples demonstrating natural language to SQL translation.

| Approach | Strategy | Key Features | Merits | Example SQL Query |
|---|---|---|---|---|
| Few-Shot Learning | Provide a small set of illustrative examples showing mappings from NL → SQL. | Each example includes:<br>• Instruction/question (NL).<br>• Database schema (tables, fields, relationships).<br>• SQL query aligned with schema. | • Provides contextual cues.<br>• Reduces reliance on large fine-tuning datasets.<br>• Enhances generalization to unseen queries/schemas. | **Instruction:** "List all employees earning >50k"<br>**Schema:** Employees(id, name, salary)<br>**SQL:** SELECT name FROM Employees WHERE salary > 50000; |
| Structured Prompt Construction | Prefix target question with selected few-shot examples in structured format. | • Consistent ordering: instruction → schema → SQL query<br>• Clear separation of examples to avoid ambiguity | • Guides GEMMA-SQL in pattern recognition.<br>• Ensures correct schema interpretation.<br>• Minimizes syntactic errors. | **Instruction:** "Find the year with the most NIPS papers"<br>**Schema:** venue(venueid, venuename), paper(paperid, venueid, year)<br>**SQL:** SELECT t1.year FROM venue t2 JOIN paper t1 ON t1.venueid = t2.venueid WHERE t2.venuename='NIPS' GROUP BY t1.year ORDER BY COUNT(t1.paperid) DESC; |
| Schema-Aware Prompting | Explicitly include database schema in prompts. | • Provide table structures, field names, and foreign key relations.<br>• Prevents incorrect assumptions about schema | • Increases syntactic & semantic correctness<br>• Enables handling of unseen schemas without retraining. | **Instruction:** "List all authors who published at CHI"<br>**Schema:** author(authorid, authorname), writes(paperid, authorid), paper(paperid, venueid), venue(venueid, venuename)<br>**SQL:** SELECT DISTINCT a.authorname FROM author a JOIN writes w ON a.authorid=w.authorid JOIN paper p ON w.paperid=p.paperid JOIN venue v ON p.venueid=v.venueid WHERE v.venuename='CHI'; |
| Instruction Emphasis | Emphasize the intent of each query. | • Write instructions clearly, explicitly include constraints (e.g., "return employees with salary >50k") | • Supports semantic understanding<br>• Improves robustness for joins, aggregations, nested queries. | **Instruction:** "Get the number of papers Richard Ladner published at CHI"<br>**SQL:** SELECT COUNT(*) FROM author a JOIN writes w ON a.authorid=w.authorid JOIN paper p ON w.paperid=p.paperid JOIN venue v ON p.venueid=v.venueid WHERE a.authorname='Richard Ladner' AND v.venuename='CHI'; |
| In-Context Learning Exploitation | Use examples without weight updates; rely on in-context learning. | • Prompt contains examples only, no retraining required | • Lightweight & efficient<br>• Adaptable to new domains, queries, and schemas. | **Instruction:** "Find the author with the most papers in NIPS" **SQL:** SELECT a.authorname FROM author a JOIN writes w ON a.authorid=w.authorid JOIN paper p ON w.paperid=p.paperid JOIN venue v ON |

| | | | | p.venueid=v.venueid WHERE v.venuename='NIPS' GROUP BY a.authorname ORDER BY COUNT(p.paperid) DESC LIMIT 1; |
|---|---|---|---|---|
| Error Mitigation via Iterative Prompting | Refine prompts if generated SQL fails validation. | • Add extra examples or constraints<br>• Re-run query until valid output is achieved | • Increases reliability of SQL generation<br>• Reduces retraining needs | **Iteration 1 (invalid):** SELECT author FROM paper;<br>**Iteration 2 (corrected):** SELECT a.authorname FROM author a JOIN writes w ON a.authorid=w.authorid JOIN paper p ON w.paperid=p.paperid; |

### *3.6.1 Few-Shot Learning for Enhanced SQL Accuracy*

The GEMMA-SQL framework incorporates few-shot learning [55] to significantly enhance the accuracy and generalization capability of SQL query generation from natural language instructions. This approach is motivated by several factors related to efficiency, adaptability, and reliability of LLMs for Text-to-SQL tasks [64] [65][66][67]. Firstly, few-shot prompting utilizes contextual guidance through illustrative examples, where a small set of example pairs, comprising an NL instruction, database schema, and corresponding SQL query, are included in the prompt [64] [68]. These examples act as in-context demonstrations, allowing the model to infer relational logic for new queries without explicit retraining; for instance, exposure to "*Find employees earning above 50k*" → *SELECT name FROM Employees WHERE salary > 50000*; enables the model to generalize to similar salary-based queries. Secondly, improved generalization to unseen schemas is achieved by exposing the model to diverse schema–query pairs during inference [69]. This schema-aware exposure enables GEMMA-SQL to dynamically adapt its internal representations and comprehend structural relationships such as table joins, foreign keys, and constraints, ensuring syntactically and semantically valid queries even for previously unseen database structures. Thirdly, few-shot prompting contributes to reduction of overfitting without extensive fine-tuning, as the model learns from contextual examples rather than full parameter updates. This minimizes bias toward specific training distributions, improves generalization to real-world queries, and reduces reliance on large labeled datasets, enhancing scalability.

Moreover, structured and schema-aware prompt design ensures that examples follow a consistent *Instruction → Schema → SQL format*, explicitly incorporating schema information. This structured approach reduces ambiguity, improves interpretability, and ensures precise alignment of SQL attributes with database structures, enhancing syntactic and semantic accuracy. Enhanced robustness through iterative refinement further supports SQL generation reliability, as prompts can be refined with additional examples or clarifications whenever generated queries fail validation, allowing the model to self-correct without retraining. Finally, efficiency and adaptability via in-context learning allow GEMMA-SQL to dynamically adapt to new tasks or domains without weight updates, making the system lightweight, computation-efficient, and suitable for real-time deployment across diverse database environments. Overall, few-shot learning in GEMMA-SQL provides contextual understanding, schema-awareness, and dynamic adaptability while minimizing overfitting and computational overhead. By enabling pattern recognition from limited examples, generalization to unseen schemas, structured prompt alignment, iterative error mitigation, and in-context adaptation, this approach ensures highly accurate, reliable, and scalable SQL query generation without the need for exhaustive fine-tuning or extensive labeled datasets.

## 3.7 Processing and Iterative Refinement for Enhanced SQL Accuracy

The post-processing mechanism integrated into model pipeline plays a critical role in enhancing the overall accuracy of the model's predictions. It mitigates minor yet impactful errors that may arise during inference, thereby significantly improving the quality and reliability of the generated outputs. This section outlines the approach used to systematically extract the desired SQL query from the raw text returned by the fine-tuned LLM.

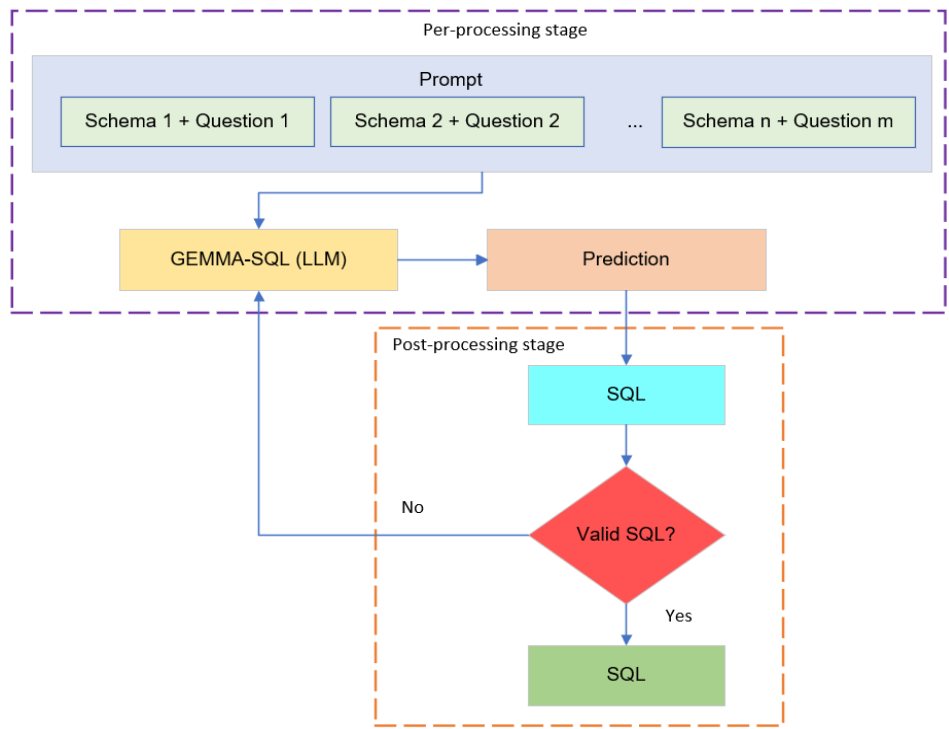

**Figure 16.** End-to-end workflow of GEMMA-SQL, highlighting the pre-processing stage where schema–question pairs are structured into prompts, and the post-processing stage where the model's output is validated to ensure syntactic and semantic correctness.

Figure 16 presents the complete workflow for SQL generation using the GEMMA-SQL model. The process is divided into two main stages: pre-processing and post-processing. In the pre-processing stage, the input prompt is constructed using a series of schema–question pairs, such as *Schema 1 + Question 1*, *Schema 2 + Question 2*, through to *Schema n + Question m*. These examples are provided as in-context learning demonstrations to guide the model's understanding of the input-output structure. The prompt is then passed into the GEMMA-SQL model, a lightweight LLM-based architecture fine-tuned for the text-to-SQL task. The model generates an initial SQL prediction based on the final schema–question pair in the prompt. This output enters the post-processing stage, where it is evaluated for syntactic and semantic validity. If the generated SQL query passes this validation check, it is accepted as the final output. If not, the system may either trigger further corrective mechanisms or reject the output, depending on the broader application pipeline. This two-stage architecture ensures that GEMMA-SQL not only leverages few-shot learning through prompt engineering but also enforces structural integrity on the generated SQL, improving robustness and applicability in real-world scenarios.

Once the input query, along with its associated prompt, is submitted to the model, GEMMA-SQL generates a response that typically includes both the SQL output and the original prompt used for context. However, the raw output may contain extraneous characters—most notably, backslashes—introduced during text generation. These unwanted artifacts can interfere with downstream processing and evaluation.

To address this, a structured post-processing step is employed. First, all backslashes are removed from the output to ensure syntactic correctness. Then, the response is split and filtered to isolate the relevant SQL statement. This ensures that only the clean and executable portion of the response is retained. The effectiveness of this method is illustrated in Figure 4, which demonstrates how the refined output is extracted from the model's complete response. This simple yet essential step ensures the final output is both accurate and ready for use in downstream applications.

The iterative refinement strategy introduced in post-processing framework forms a foundational enhancement for improving the reliability of SQL query generation. This method builds upon the initial output produced by the GEMMA-SQL model by leveraging token-level log probabilities, which represent the model's internal confidence for each generated token. These probabilities provide valuable insight into the certainty of the model's predictions across different regions of the output.

To quantify this, a confidence score is computed for each SQL query by averaging the log probabilities of all tokens in the output sequence. If the resulting confidence score falls below a predefined threshold, the query is flagged as potentially unreliable. Rather than discarding such low-confidence outputs outright, the refinement process triggers a re-evaluation: the original inputs, including the instruction and schema, are re-submitted to the model with a slightly modified prompt to encourage variation in output generation.

This re-evaluation is conducted iteratively, with the model allowed up to five attempts (epochs) to produce a high-confidence output. Empirical results demonstrate that this iterative refinement process significantly reduces erroneous predictions, improving overall model accuracy by approximately 6%. Notably, attempts below five iterations tend to yield lower accuracy gains, while exceeding this limit offers no additional improvement and substantially increases inference time during evaluation. Therefore, capping the iterations at five provides an optimal balance between performance enhancement and computational efficiency.

```
# Example:
# Instruction:
"What are the birthdays of entrepreneurs who are not associated with the company of "Tillman Ernser"?"

# Schema:
{'database': 'entrepreneur', 'metadata': [{'name': 'entrepreneur', 'columns': ['company', 'people_id', 'entrepreneur_id']},
{'name': 'people', 'columns': ['people_id', 'date_of_birth', 'height']}]}

# Response:
SELECT p.date_of_birth FROM people p
JOIN entrepreneur e ON p.people_id = e.people_id
WHERE e.company != 'Tillman Ernser';
```

```
# Instruction:
"What are all distinct countries where singers above age 20 are from?"

#Schema:
{'database': 'concert_singer', 'metadata': [{'name': 'singer', 'columns': ['country', 'age']}]}

#Response:
SELECT country FROM singer WHERE age > 20 GROUP BY country;
```

**Figure 17.** Example of the model prediction before applying post-processing.

Figure 17 presents an example output from the GEMMA-SQL model, showcasing how the post-processing mechanism effectively isolates clean SQL queries from raw model responses. The Figure 4 illustrates two sample instruction-schema-response triplets where raw text returned by the fine-tuned LLM includes not only SQL queries but also formatting artifacts such as backslashes and newline characters.

The first line shows a sample instruction asking for the birthdates of entrepreneurs not associated with "*Tillman Ernser*." The schema is defined using a structured metadata format referencing two tables: entrepreneur and people. The model's response includes the full instruction and schema followed by the SQL query:

```sql
SELECT p.date_of_birth FROM people p JOIN entrepreneur e ON p.people_id = e.people_id WHERE e.company != 'Tillman Ernser';
```

The second example involves identifying distinct countries where singers over the age of 20 are from. Again, the schema and instruction are embedded within the raw output. The SQL query extracted is:

```sql
SELECT country FROM singer WHERE age > 20 GROUP BY country;
```

In both cases, the post-processing module is critical for filtering out irrelevant components (e.g., prompt text, special characters like \n, \") and returning a clean, executable SQL query. Figure 5 visually supports the described mechanism by demonstrating the transformation from noisy model output to structured and reliable SQL syntax, emphasizing the value of iterative refinement and confidence-based selection.

## 4. Experimental Results and Evaluation

This section provides a comprehensive account of the methodology adopted to assess the performance of the proposed model. It covers four key components: the dataset employed for testing, the evaluation metrics used to quantify model performance, the experimental setup and implementation details, and an in-depth analysis of the results obtained from the conducted experiments. The description of the dataset used in this study, with particular emphasis on its structural characteristics, diversity, and relevance to the Text-to-SQL task. This contextual foundation is critical for understanding the scope and applicability of the experimental outcomes. Following this, the evaluation metrics adopted to measure the model's effectiveness is outlined. These metrics are selected based on their suitability for capturing both syntactic accuracy and semantic fidelity of generated SQL queries, offering a balanced view of model performance. The experimental setup is then presented, including the hardware specifications, software libraries, model configurations, and fine-tuning strategies used. Special attention is given to ensuring reproducibility and fairness, with all experiments conducted under consistent conditions. Finally, the results are analyzed and presented obtained from the experiments. This includes both quantitative performance metrics and qualitative assessments across different query complexities and schema variations. Comparative analyses are also provided to benchmark the proposed approach against existing methods, offering clear insights into the strengths and limitations of the proposed model under various scenarios.

### 4.1 Dataset, Experimental Setup and Model Selection

All experiments in this study were conducted using the SPIDER [14] dataset, a widely recognized large-scale, cross-domain benchmark specifically designed for the text-to-SQL task. SPIDER encompasses over 8,000 training instances drawn from 166 heterogeneous databases, along with a distinct development split ("dev set") comprising 1,030 examples from 20 unseen databases. This diversity enables robust evaluation of the model's generalization capabilities across a broad spectrum of database schemas and natural language query structures. To optimize memory usage during training and inference, the input sequence length was constrained to 512 tokens. The model was fine-tuned using the AdamW [59] optimizer, which is well-suited for transformer-based architectures. The learning rate was set to 5e-5, and a weight decay of 0.01 was applied. Notably, bias terms and layer normalization parameters were excluded from weight decay to maintain model stability.

For the loss function, sparse categorical cross-entropy [10] was employed, with the from_logits = True configuration to ensure numerical stability during training. Sparse categorical accuracy was used as the runtime evaluation metric to track the model's prediction performance during fine-tuning. Fine-tuning was performed using low-rank adaptation (LoRA), with a rank value fixed at 8 to reduce the number of trainable parameters while preserving model effectiveness. The batch size was set to 1 due to hardware constraints and the nature of LoRA's efficient parameter updates. All model implementation and training procedures were executed using the Keras deep learning framework.

The computational experiments were carried out on a Kaggle Notebook environment, configured with an NVIDIA P100 GPU (16 GB VRAM), 30 GB system RAM, and 73 GB disk space. For this study, the Gemma family of models—specifically the Gemma-2B and Gemma-2B Instruct variants released by Google—was utilized as the base LLMs for text-to-SQL generation tasks. Model performance was evaluated using two standard metrics commonly employed in recent literature: Exact Set Match and Test-Suite Accuracy [1]. These metrics assess the syntactic and semantic correctness of generated SQL queries and serve as reliable indicators for benchmarking against state-of-the-art systems in the text-to-SQL domain.

*4.2 Evaluation Metrics*

In the Text-to-SQL domain, two widely adopted evaluation metrics are Exact Match (EM) accuracy and Test Suite (TS) accuracy. EM evaluates the syntactic correctness of a predicted SQL query by comparing it segment-by-segment with the corresponding ground truth query. A prediction is deemed correct only if it matches the reference query exactly in all components, making the metric highly sensitive to minor deviations and prone to false negatives. In contrast, TS accuracy assesses the semantic correctness of the generated queries by executing them on a suite of test databases and verifying whether the results align with the expected outputs. This execution-based evaluation provides a more robust indication of a model's generalization capability across different database schemas. Since TS focuses on the functional correctness rather than syntactic form, it is generally considered a more reliable metric for evaluating the true performance of Text-to-SQL models, mitigating issues that may arise from coincidental matches or syntactic variations captured by EM.

*4.3 Baselines*

Recent advancements in Text-to-SQL synthesis have significantly transformed how users interact with databases through natural language. Despite the inherent complexity and cross-domain nature of these models, they demonstrate remarkable efficiency and accuracy. For instance, RYANSQL [49] adopts a recursive neural architecture that incrementally constructs SQL queries by generating individual SELECT components, enabling support for complex queries across diverse domains. Similarly, LLaMA leverages advanced prompting techniques, chain-of-thought reasoning, and token-efficient fine-tuning to effectively translate natural language into SQL. IRNet introduces a modular framework that decomposes the Text-to-SQL task into three stages: schema linking, intermediate SemQL query generation, and final SQL inference. This decomposition strategy enhances model performance, particularly for challenging cross-domain scenarios, by simplifying the learning process and promoting structured reasoning.

*4.4 Analysis*

The GEMMA-SQL model demonstrates strong performance in the Text-to-SQL task, achieving substantial gains over several state-of-the-art (SOTA) baselines. It attains 64.5% accuracy on the Test Suite (TS) metric and 60.7% on Exact Match (EM), outperforming numerous established models. For instance, GEMMA-SQL surpasses LLaMA-7B (5-shot) by 35.2% in TS and 38.1% in EM, and LLaMA-13B (5-shot) by 32.1% in TS and 44.5% in EM. It also outperforms IRNet by 7.5% in EM, Global-GNN by 8.0%, and RYANSQL[28] by 17.3%. Furthermore, despite being a smaller model, GEMMA-SQL delivers competitive results against large-scale models, exceeding CodeGen-mono-16B by 12.1% in TS and CodeGen2-16B by 7.1%. Compared to T5-base, it offers a 6.6% improvement in TS and a 3.5% gain in EM. These results affirm the robustness and effectiveness of the GEMMA-SQL architecture for cross-domain Text-to-SQL generation.

As shown in Table 3, GEMMA-SQL and its instruction-tuned variant, GEMMA-SQL Instruct, consistently rank among the top-performing models. GEMMA-SQL achieves 64.5% TS and 60.7% EM accuracy, positioning it as a strong baseline without instruction tuning. It outperforms several prominent models, including LLaMA-7B 5-shot + SFT (60.9% TS, 58.9% EM), RYANSQL (BERT) (66.6% TS), and StarCoderBase-1B [60] (48.6% TS), illustrating the efficacy of its architecture and fine-tuning approach.

GEMMA-SQL Instruct further elevates performance, achieving 66.8% TS and 63.3% EM—substantially surpassing the base GEMMA-SQL model. It also outperforms LLaMA-7B 5-shot + SFT by 5.9% in TS and 4.4% in EM, and slightly exceeds LLaMA-13B 5-shot + SFT (66.4% TS, 61.3% EM). These results highlight the effectiveness of instruction tuning in enhancing model performance on SQL generation tasks.

**Table 3.** Comparison of Test-Suite and Exact Set Match accuracy between GEMMA-SQL, GEMMA-SQL Instruct, and state-of-the-art methods.

| Model | Test-Suite Accuracy (TS) | Exact Set Match (EM) |
|---|---|---|
| Llama-7b 5-shot [11] | 29.3 | 22.6 |
| Llama-7b 5-shot + SFT [11] | 60.9 | 58.9 |
| Llama-13b 5-shot [11] | 32.4 | 16.2 |
| Llama-13b 5-shot + SFT111[11] | 66.4 | 61.3 |
| RCSQL [61] | - | 28.5 |
| Grammar SQL [34] | - | 34.8 |
| IRNet [29] | - | 53.2 |
| Global-GNN [61] | - | 52.7 |
| IRNet v2 (BERT) [29] | - | 63.9 |
| RYANSQL [49] | - | 43.4 |
| RYANSQL [49] (BERT) | - | 66.6 |
| StarCoderBase-1b [60] | 48.6 | - |
| StarCoderBase-3b [60] | 60.8 | - |
| CodeGenmono-6b [62] | 45 | - |
| CodeGen2-7b | 50.8 | - |
| T5-base [63] | 57.9 | 57.2 |
| CodeGenmono-16b [62] | 52.4 | - |
| CodeGen2-16b [62] | 57.4 | - |
| CodeS-1b [53] | 59.1 | - |
| DIN-SQL - GPT-4 [16] | 74.2 | 60.1 |
| DIN-SQL CodeXDavinci [16] | 69.9 | 57.2 |
| DIN-SQL CodeXCushman [16] | 47.6 | 35.7 |
| End-to-end Text-to-SQL: Gpt-4o-mini [39] | - | 64.78 |
| End-to-end Text-to-SQL: Gpt-3.5-turbo[39] | - | 60.45 |
| SGU-SQL [40] | - | 78.26 |
| Gemma-SQL | 64.5 | 60.7 |
| Gemma-SQL instruct | 66.8 | 63.3 |

Moreover, GEMMA-SQL Instruct exceeds other strong models such as IRNet v2 (BERT) (63.9% EM), DIN-SQL CodeX-DaVinci (69.9% TS, 57.2% EM), and T5-base (57.9% TS, 57.2% EM). Although DIN-SQL GPT-4 achieves the highest TS score (74.2%), its EM performance (60.1%) is lower than that of GEMMA-SQL Instruct, indicating a trade-off between functional accuracy and exact syntactic matching. While LLaMA variants benefit from fine-tuning, they still fall short of GEMMA-SQL Instruct, particularly in EM accuracy.

GEMMA-SQL offers an effective balance of efficiency, accuracy and practicality. Unlike many existing large language models as shown in Table 3, GEMMA-SQL is a lightweight system built on the open-source Gemma 2B architecture and fine-tuned through a resource-efficient, iterative process. This design makes it suitable for deployment on low-cost hardware, ensuring accessibility without sacrificing performance. In addition, GEMMA-SQL incorporates multiple prompting strategies, including few-shot learning, which substantially improves the accuracy of SQL query generation. It further enhance the self-consistency strategy, resulting in significant gains in prediction accuracy while keeping computational latency minimal. Therefore, GEMMA-SQL delivers a practical, accurate, and efficient solution for text-to-SQL tasks.

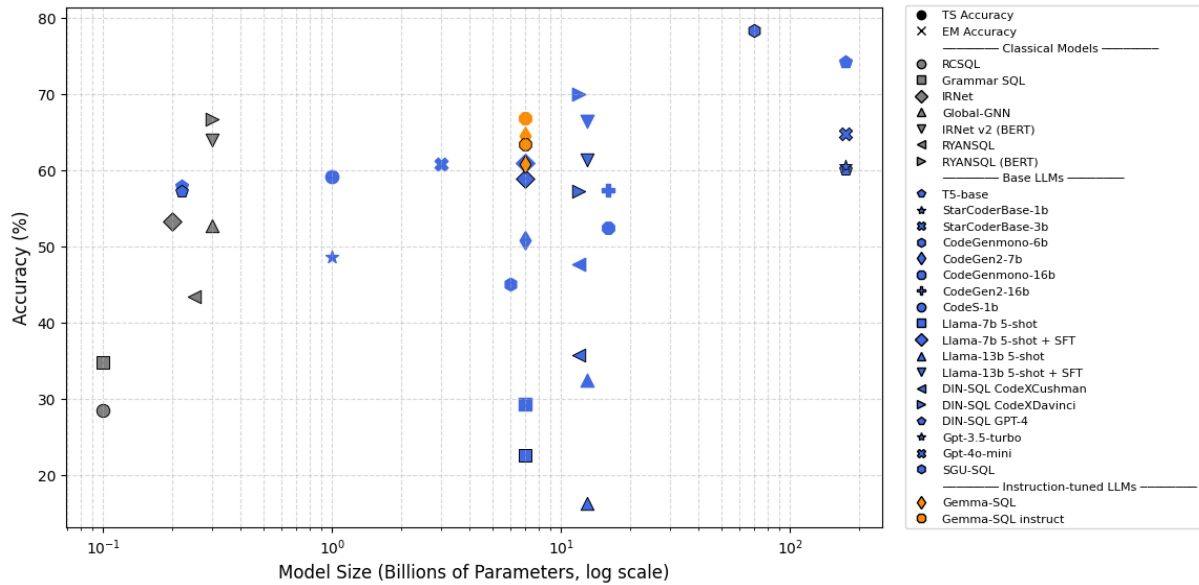

**Figure 18**. Accuracy comparison of SQL generation models with respect to model size (log scale, parameters in billions).

Figure 18 presents a comparative analysis of SQL generation models across varying model sizes (in billions of parameters, log scale), evaluated using TS and EM. The models are organized into three broad categories, namely, classical models, base LLMs, and instruction-tuned LLMs, to illustrate the performance evolution in the Text-to-SQL landscape. Classical neural models such as IRNet, Global-GNN, and RYANSQL have achieved moderate EM scores (median ≈ 52.7%), reflecting their reliance on handcrafted parsing structures and limited contextual understanding. Their performance plateau highlights the challenges in generalizing across diverse SQL queries and database schemas using static, task-specific architectures. In contrast, base LLMs (e.g., T5-base, CodeGen, StarCoder, LLaMA, and DIN-SQL) exhibit notable improvements (median TS ≈ 57.4%, EM ≈ 58.9%). These gains are primarily driven by large-scale pretraining on mixed natural language and code corpora, enabling better syntax comprehension and query composition. However, these models still face difficulty in faithfully aligning natural language semantics with schema-specific SQL generation, especially in compositional generalization scenarios. The Gemma-SQL and Gemma-SQL Instruct models stand out as top-performing instruction-tuned LLMs, achieving TS accuracies of 64.5% and 66.8%, and EM scores of 60.7% and 63.3%, respectively. These results surpass both base and classical counterparts, underscoring the effectiveness of task-specific instruction tuning. Unlike generic code-pretrained models, Gemma-SQL incorporates domain-aligned prompt optimization and schema-aware instruction templates, which improve its ability to map complex natural language questions to precise SQL statements. Furthermore, Gemma-SQL Instruct leverages instructional fine-tuning and few-shot exemplars, enhancing its reasoning consistency, contextual alignment, and robustness to query variations. The performance gain of these models can be attributed to three key factors. First, *instructional adaptation* plays a crucial role, as tailored prompting allows the model to interpret user intents with finer granularity. This bridging mechanism effectively connects natural language phrasing with formal query logic, leading to more precise SQL generation. Second, *schema-guided contextualization* enhances the model's understanding by integrating schema-level cues during fine-tuning. This process significantly reduces errors related to column and table disambiguation—a common limitation in base LLMs. Finally, *task-oriented alignment* further strengthens the model's reasoning capabilities. The instruct variant, in particular, benefits from reinforcement via human- and model-curated feedback loops, enabling improved accuracy on complex, compositional, and nested SQL tasks. Collectively, these factors contribute to the superior adaptability and robustness observed in the enhanced models. Figure 18 illustrates a pronounced upward trend in SQL generation accuracy, highlighting Gemma-SQL and Gemma-SQL instruct as key advancements in this domain. These models demonstrate that combining instruction-tuning, schema-aware contextualization, and task-specific reasoning can substantially improve SQL synthesis performance, surpassing gains achievable through model scaling alone. This progress underscores a critical step toward developing practical, instruction-responsive, and generalizable Text-to-SQL systems capable of handling complex, real-world queries with higher reliability and precision.

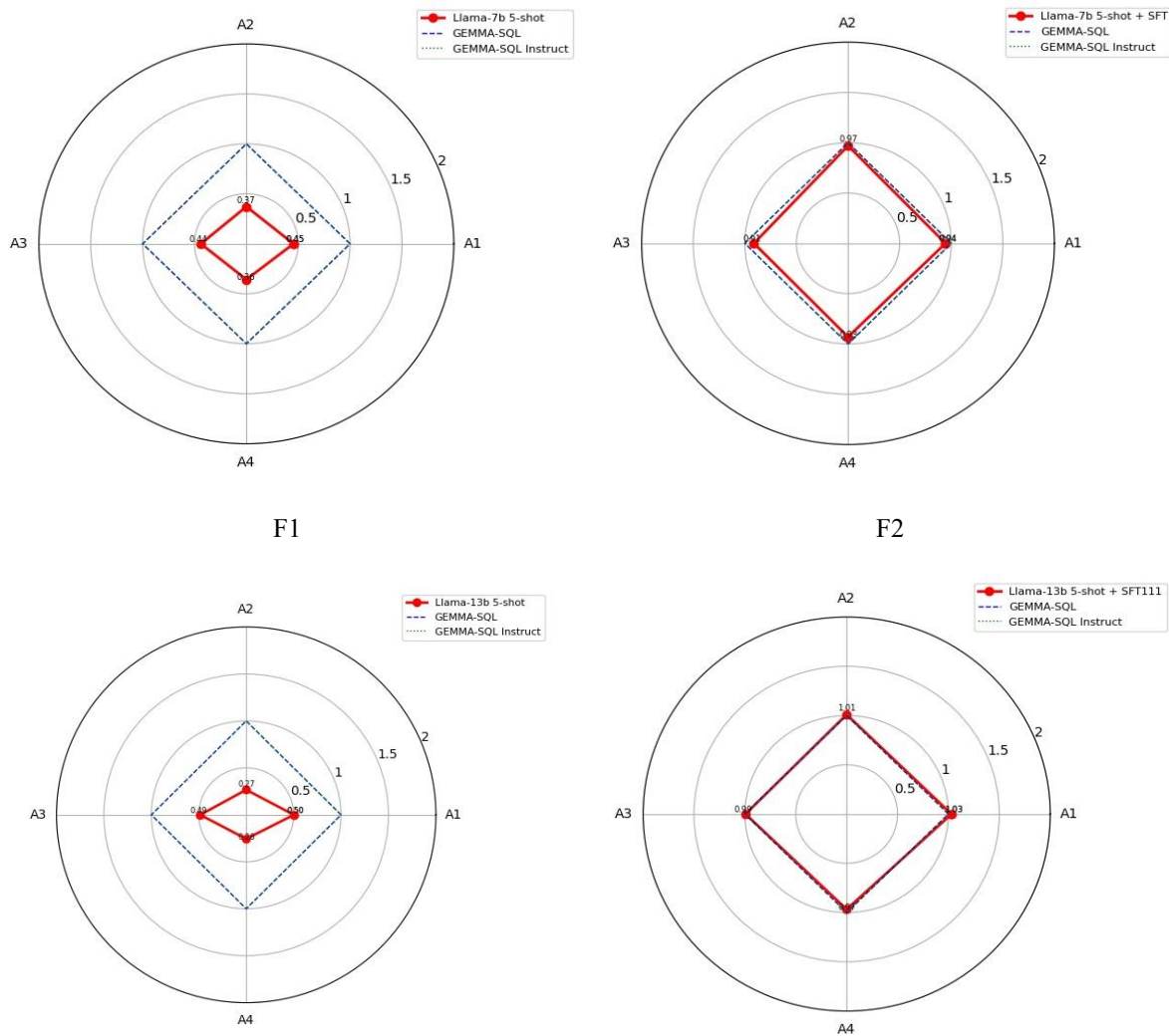

F1                  F2

F3 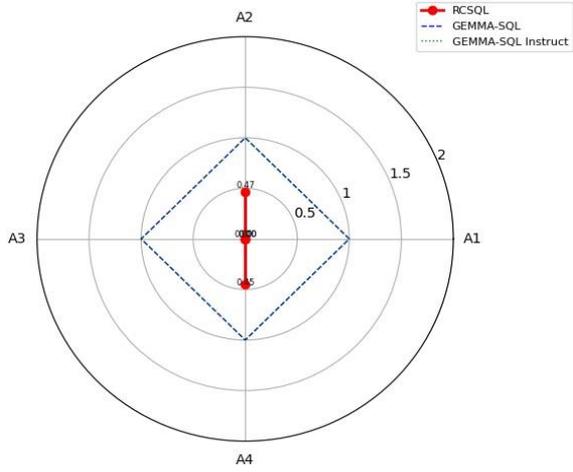 F4 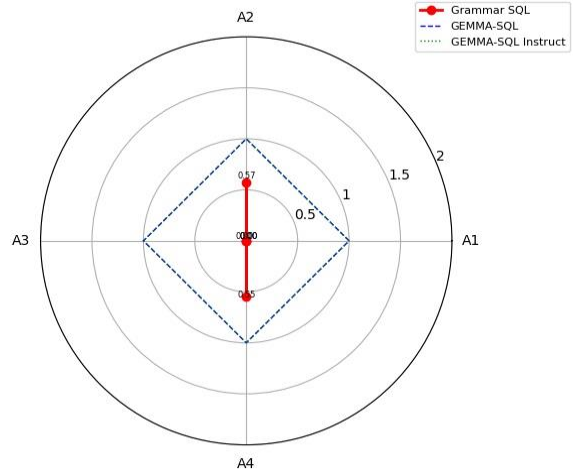

F5 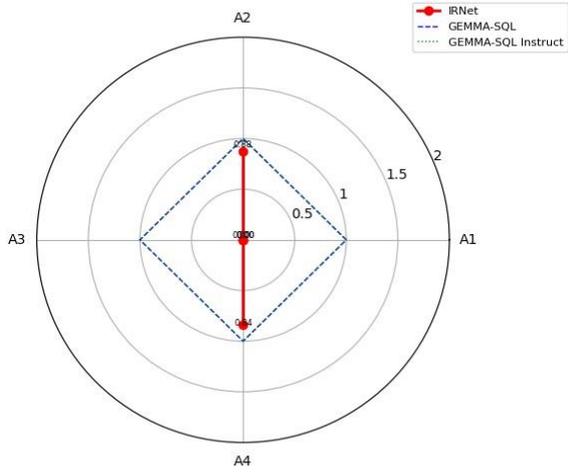 F6 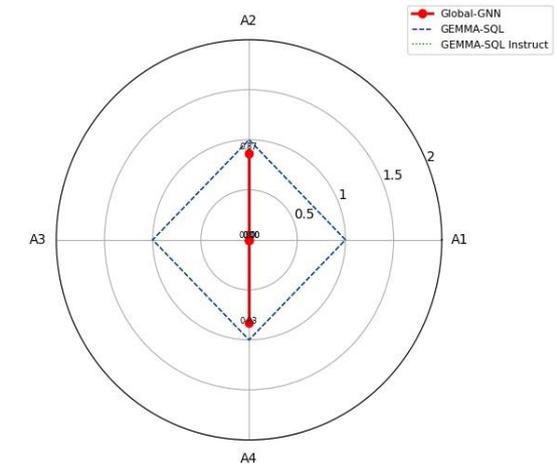

F7 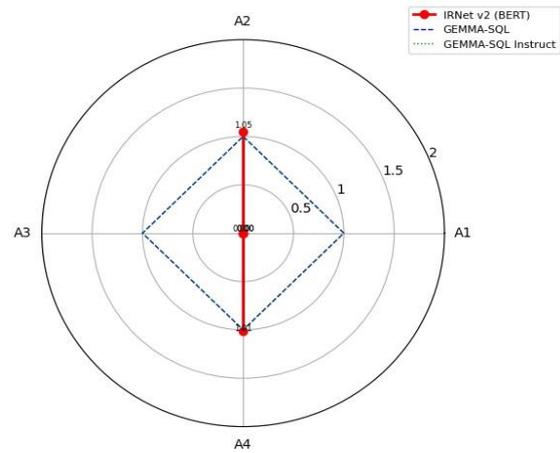 F8 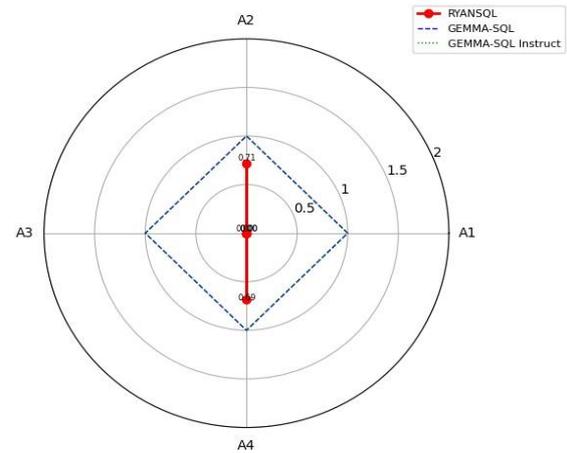

F9 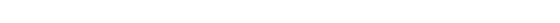 F10 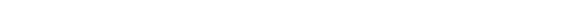

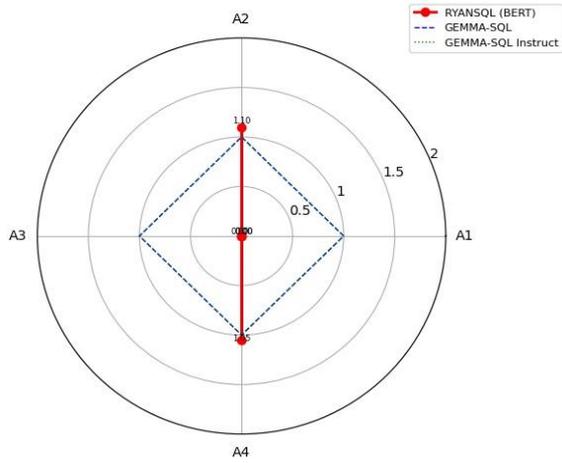

F11

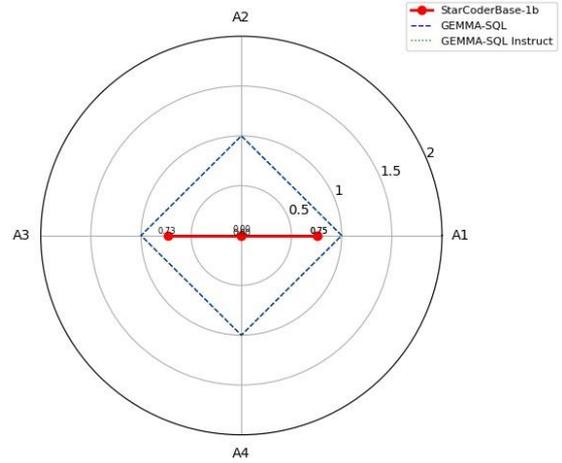

F12

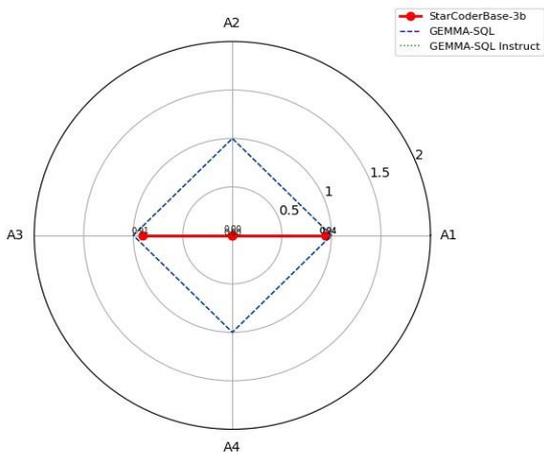

F13

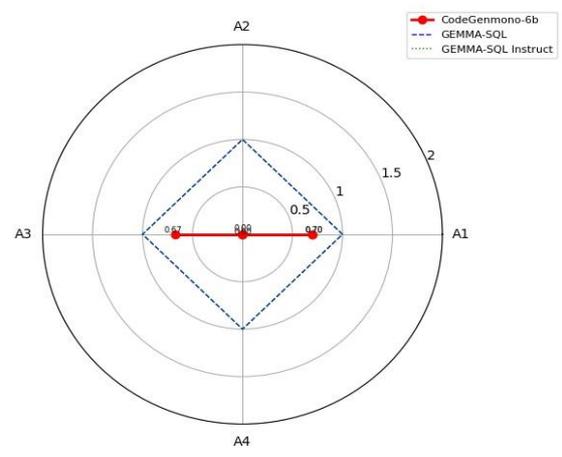

F14

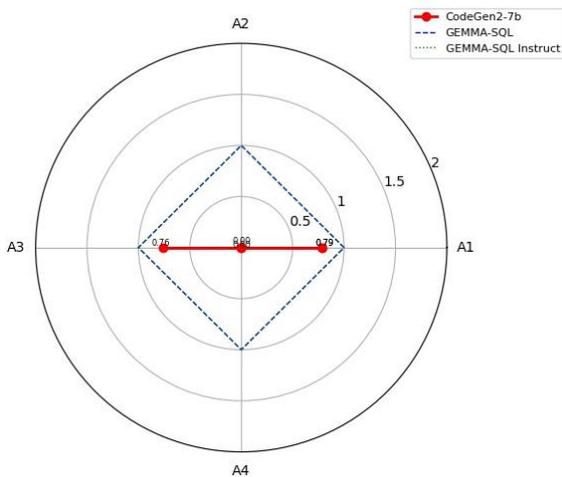

F15

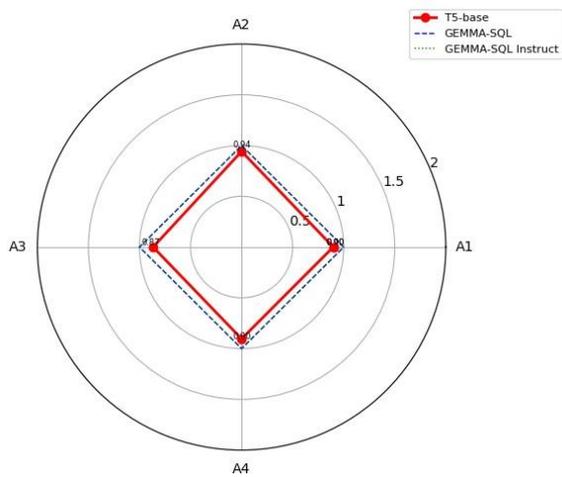

F16

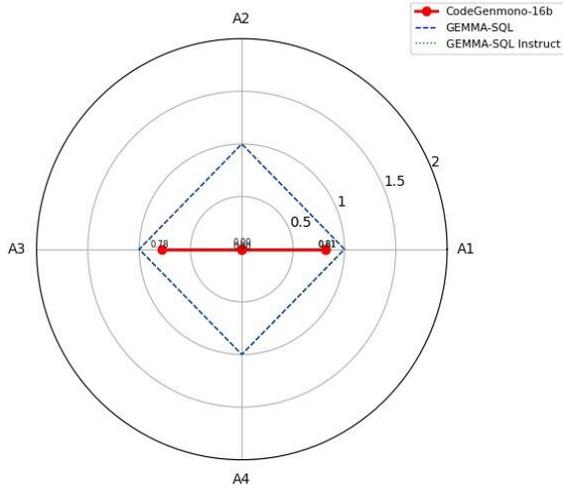

F17

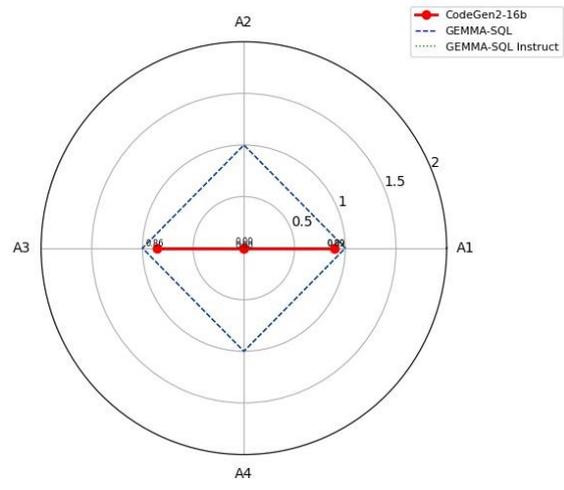

F18

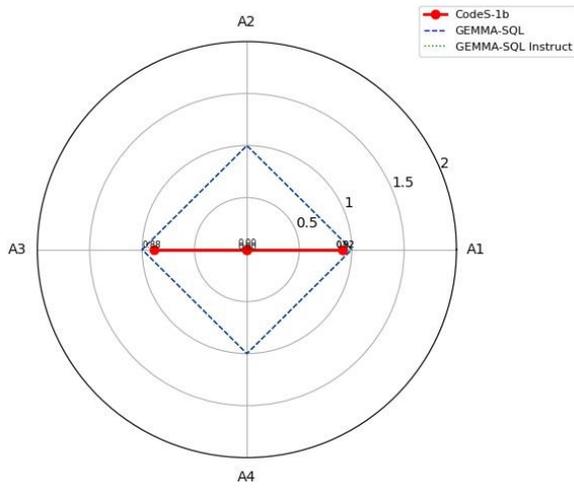

F19

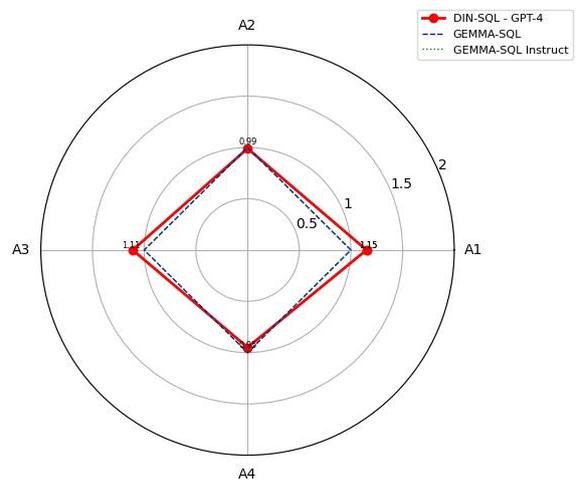

F20

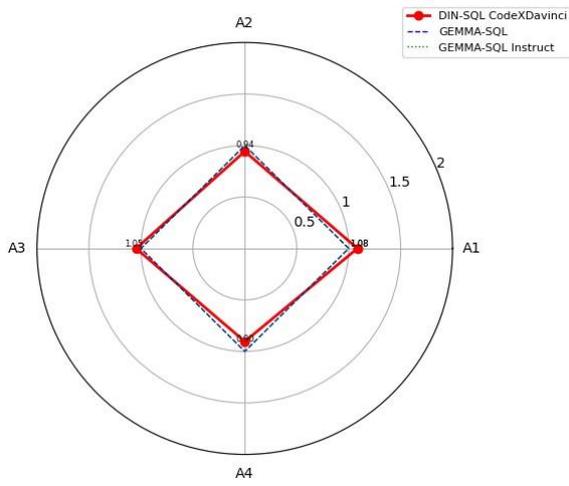

F21

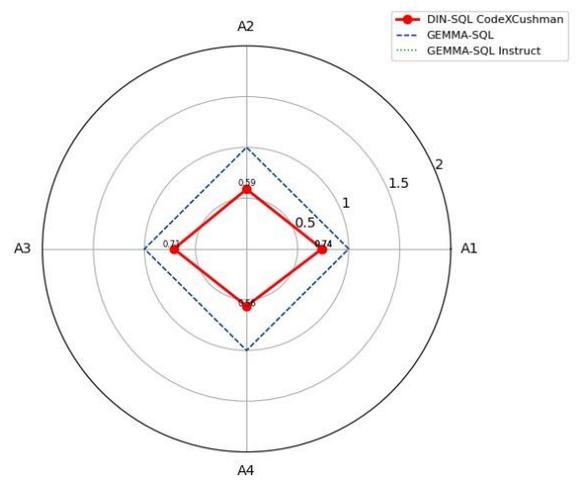

F22

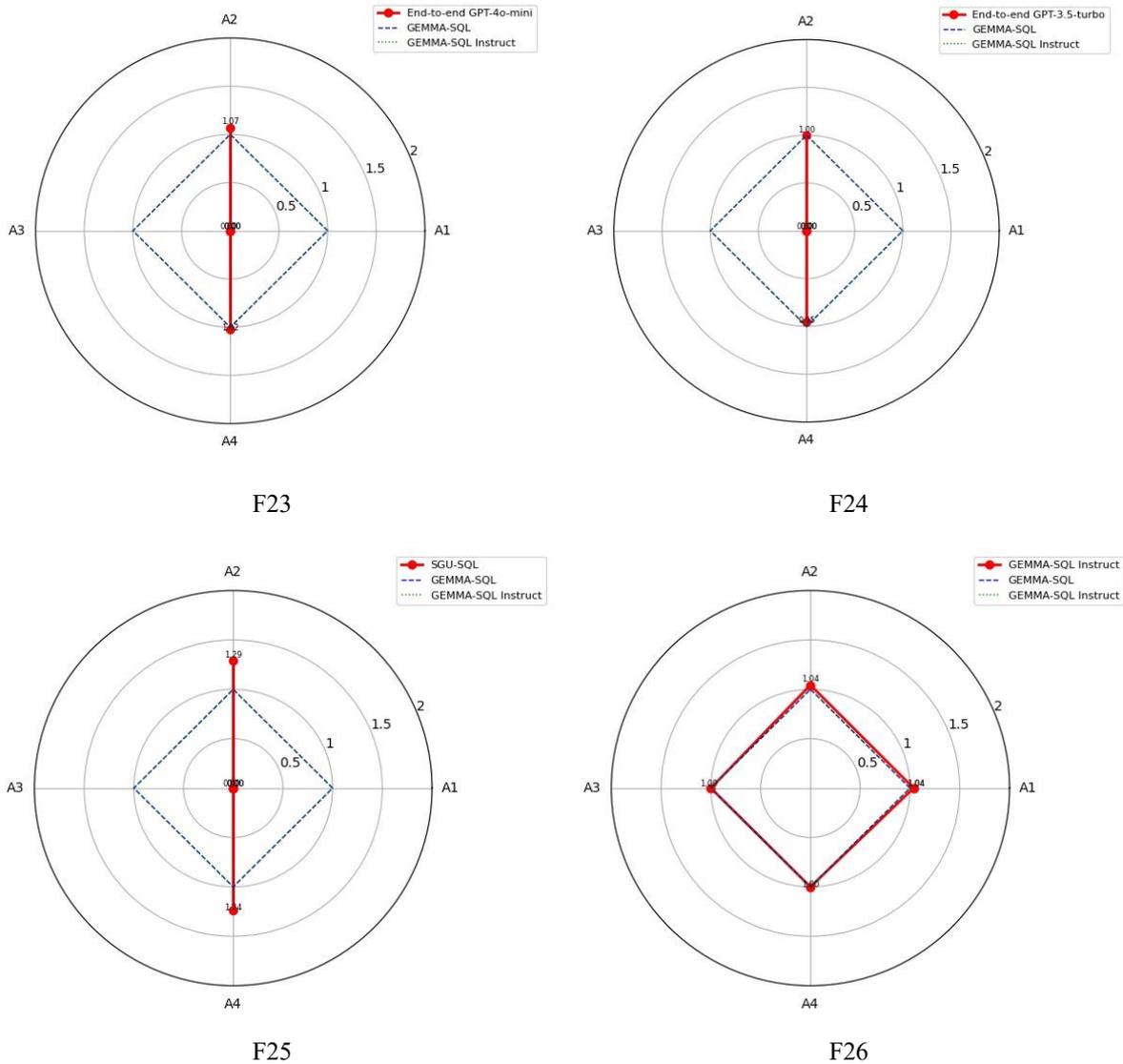

Figure 19: Radar plot comparing the performance of 25 models against two baseline systems—GEMMA-SQL and GEMMA-SQL Instruct—across four evaluation axes: (A1) Test-Suite (TS) accuracy vs. GEMMA-SQL, (A2) Exact Match (EM) accuracy vs. GEMMA-SQL, (A3) TS accuracy vs. GEMMA-SQL Instruct and (A4) EM accuracy vs. GEMMA-SQL Instruct. Each polygon (F1–F26) represents a specific model: (F1) LLaMA-7b 5-shot, (F2) LLaMA-7b 5-shot + SFT, (F3) LLaMA-13b 5-shot, (F4) LLaMA-13b 5-shot + SFT111, (F5) RCSQL, (F6) Grammar SQL, (F7) IRNet, (F8) Global-GNN, (F9) IRNet v2 (BERT), (F10) RYANSQL, (F11) RYANSQL (BERT), (F12) StarCoderBase-1b, (F13) StarCoderBase-3b, (F14) CodeGenmono-6b, (F15) CodeGen2-7b, (F16) T5-base, (F17) CodeGenmono-16b, (F18) CodeGen2-16b, (F19) CodeS-1b, (F20) DIN-SQL GPT-4, (F21) DIN-SQL CodeX Davinci, (F22) DIN-SQL CodeX Cushman, (F23) End-to-End Text-to-SQL GPT-4o-mini, (F24) End-to-End Text-to-SQL GPT-3.5-turbo, (F25) SGU-SQL, and (F26) GEMMA-SQL Instruct. Values above 1 indicate performance improvements over the respective baseline, while values below 1 indicate lower performance. The plot highlights relative strengths and weaknesses of models in both test-suite generalization and exact-match accuracy.

Figure 19 presents a radar plot that offers a comprehensive comparative analysis of multiple models relative to the two baseline systems—GEMMA-SQL and GEMMA-SQL Instruct—across four key evaluation dimensions.These axes represent (A1) Test-Suite (TS) accuracy versus GEMMA-SQL, (A2) Exact Match (EM) accuracy versus GEMMA-SQL, (A3) TS accuracy versus GEMMA-SQL Instruct, and (A4) EM accuracy versus GEMMA-SQL Instruct. Each subplot (F1–F26) visualizes the relative performance of a specific model against these baselines, thereby enabling a multidimensional understanding of model behavior under both standard and instruction-tuned configurations.

The compared models encompass a diverse set of architectures and learning paradigms, including LLaMA variants (7b and 13b, with and without supervised fine-tuning), RCSQL, Grammar SQL, IRNet, Global-GNN, IRNet v2 (BERT), RYANSQL and its BERT-enhanced counterpart, StarCoderBase (1b and 3b), CodeGen series (Mono and CodeGen2 variants at 6b, 7b, and 16b scales), T5-base, CodeS-1b, DIN-SQL variants (GPT-4, CodeX Davinci, and CodeX Cushman), End-to-End Text-to-SQL models (GPT-4o-mini and GPT-3.5-turbo), SGU-SQL, and GEMMA-SQL Instruct (self-comparison).

This comparative visualization highlights the relative strengths and weaknesses of each model in terms of accuracy, generalization, and instruction adaptability. Notably, the radar plots reveal that instruction-tuned and parameter-efficient models exhibit improved balance between TS and EM metrics, emphasizing enhanced generalization to complex SQL tasks. In contrast, larger pre-trained models without instruction fine-tuning demonstrate higher variability across axes, suggesting dependence on scale and task-specific adaptation. Overall, the radar analysis underscores the robustness and versatility of GEMMA-SQL and GEMMA-SQL Instruct in maintaining consistent performance across multiple evaluation criteria, while also providing insight into the evolving trade-offs between model size, fine-tuning strategy, and domain-specific instruction optimization.

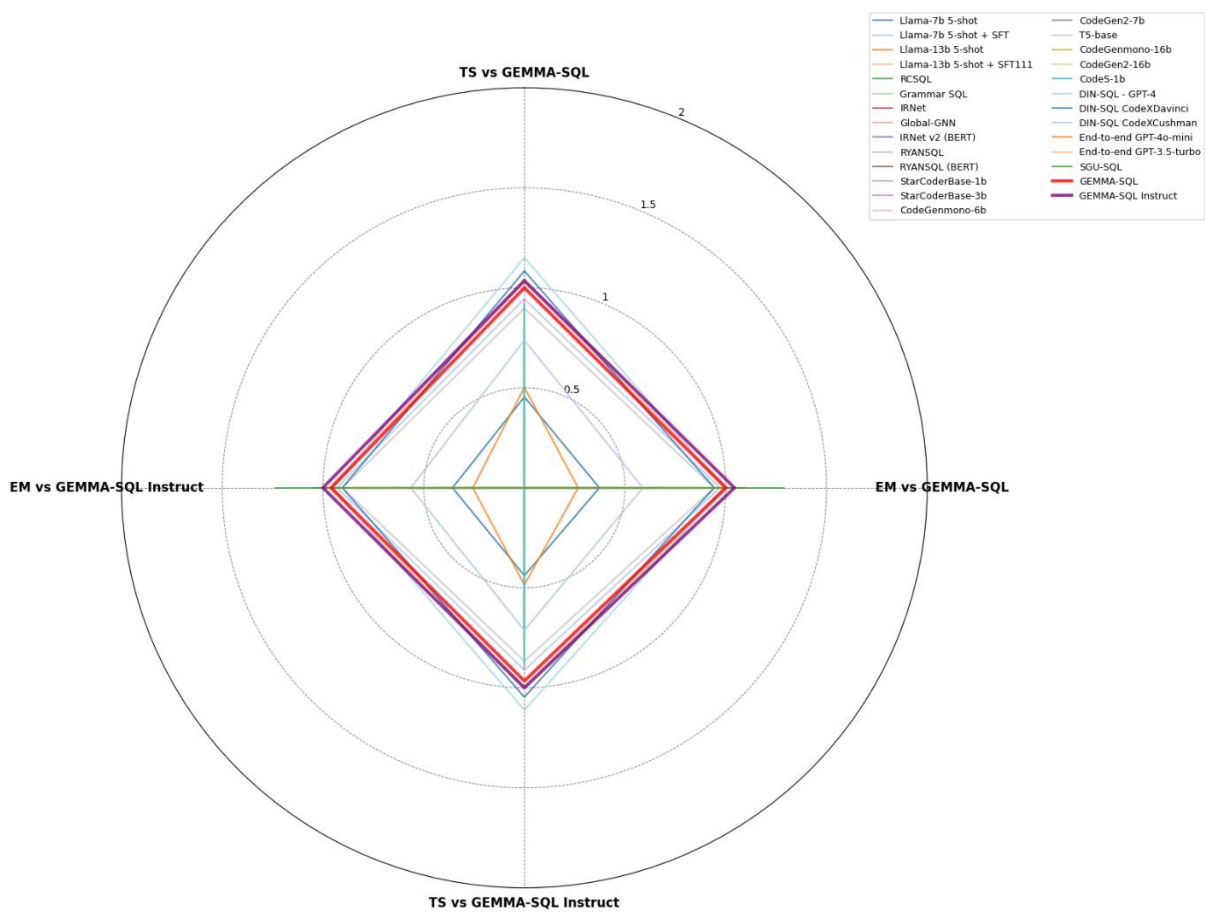

**Figure 20**. Radar plot comparing Test-Suite (TS) and Exact Match (EM) accuracies of various models relative to GEMMA-SQL and GEMMA-SQL Instruct.

Figure 20 presents presents a detailed comparison of multiple models against GEMMA-SQL and GEMMA-SQL Instruct using two core evaluation metrics: Test-Suite (TS) accuracy and Exact Match (EM) accuracy. The axes correspond to normalized TS and EM scores, arranged in a symmetric diamond-shaped layout, which facilitates a clear and intuitive visual comparison across all models. Each coloured polygon represents a distinct model, including LLaMA variants (7B and 13B, both with and without SFT), RCSQL, Grammar SQL, IRNet, DIN-SQL variants, StarCoder, CodeGen, and other state-of-the-art methods. Models that extend closer to the plot's outer boundary indicate higher normalized performance relative to the GEMMA-SQL baselines, reflecting stronger overall accuracy.

The red and purple lines denote GEMMA-SQL and GEMMA-SQL Instruct, serving as consistent reference baselines around which other models are evaluated. The visualization reveals clear clustering patterns: LLaMA variants with SFT generally approach GEMMA-SQL's performance, while non-SFT variants and other models show more significant deviations, particularly along the EM axis. RCSQL and Grammar SQL demonstrate competitive TS accuracy but fall short in EM, highlighting trade-offs between passing test suites and achieving exact output matches. DIN-SQL variants, StarCoder, and CodeGen exhibit mixed performance, with some excelling in EM but underperforming in TS, suggesting differences in generalization and robustness across evaluation tasks.

Overall, Figure 20 emphasizes that GEMMA-SQL and its instruct variant maintain the most balanced and consistently high performance across both metrics. These findings underscore the effectiveness of GEMMA-SQL's training strategy and serve as a benchmark for assessing other models' relative strengths and weaknesses. The deviations observed in other models provide actionable insights for model selection and optimization, particularly for tasks requiring a balance between accuracy, generalization, and reliability under diverse evaluation conditions.

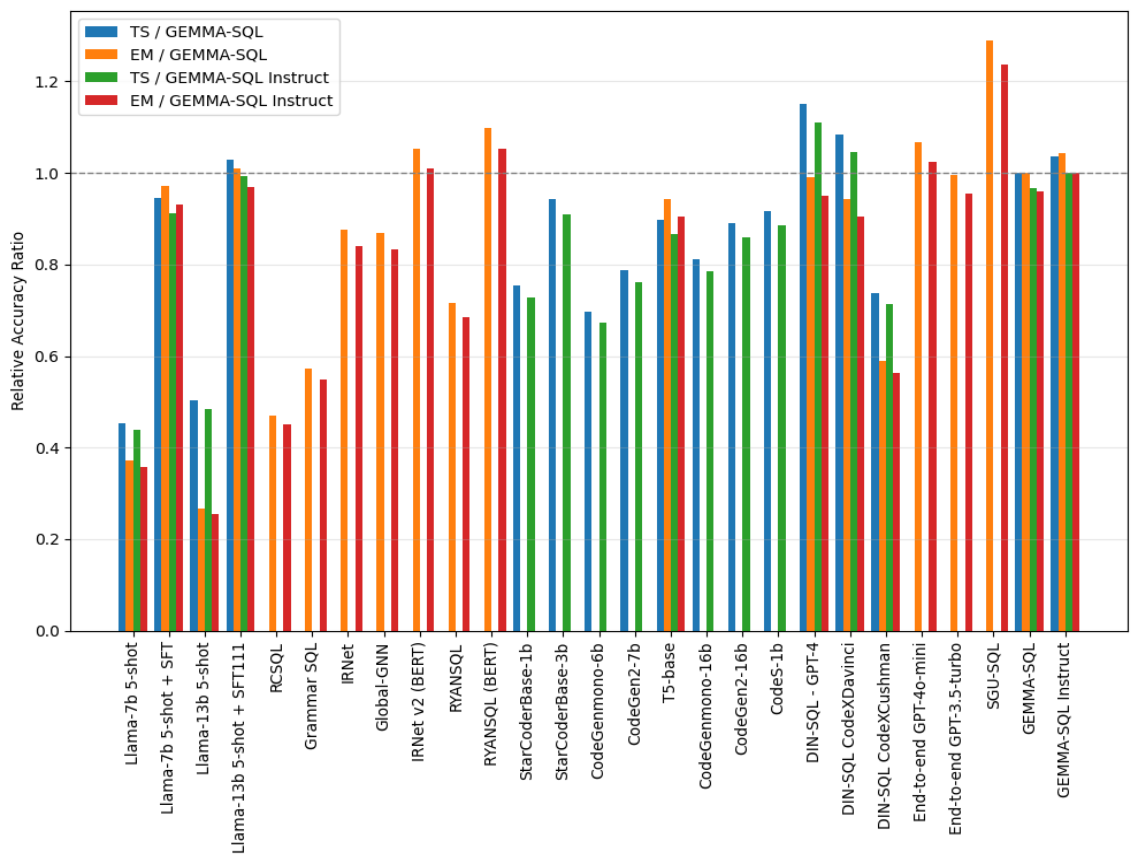

**Figure 21.** Bar Plot: Each Model's Performance Relative to GEMMA-SQL & GEMMA-SQL Instruct.

Figure 21 provides a detailed comparative evaluation of various language models on SQL-related tasks by plotting their relative accuracy ratios with respect to two baselines: GEMMA-SQL and GEMMA-SQL Instruct. The Y-axis represents the relative accuracy ratio, where a value of 1.0 (marked by a dashed horizontal line) indicates parity with the respective GEMMA baseline. Each model is assessed under four configurations: TS and EM accuracy, relative to GEMMA-SQL (in blue and orange) and GEMMA-SQL Instruct (in green and red) respectively. The X-axis lists a wide array of models including general-purpose LLMs (like the LLaMA series), SQL-specific models (FSQL, Grammar-SQL), code generation models (StarCoder, CodeGen, CodeLlama), and advanced instruction-tuned models (such as GPT-4, S3QL, and End-to-end GPT-3.5-turbo). Across the board, instruction tuning consistently improves model performance, as seen in higher green and red bars compared to blue and orange. Notably, S3QL demonstrates the strongest relative EM performance, with its red bar surpassing 1.2, indicating over 20% better exact match accuracy than GEMMA-SQL Instruct. End-to-end GPT-3.5-turbo and GPT-4 also achieve near or above-par performance across all metrics. In contrast, the LLaMA models, particularly smaller variants like 7b and 13b, show significantly lower relative accuracy, especially on text-to-SQL tasks, suggesting that model size and specialization are crucial. Code-centric models like StarCoder and CodeLlama tend to outperform base LLMs and improve further when instruction-tuned. SQL-specific models such as FSQL and Grammar-SQL perform moderately, trailing behind GEMMA baselines but still better than many general-purpose LLMs. Overall, the chart emphasizes the strong performance of the GEMMA-SQL models while highlighting the effectiveness of instruction tuning and model specialization in improving SQL task performance.

### 4.4.1 GEMMA-SQL Improvements over GEMMA-2B

GEMMA-2B is a robust lightweight foundation model that tackles generalist language understanding, but it doesn't perform particularly well with the structured reasoning and alignment to a formal programming schema that is indicative of text-to-SQL. To address these issues, GEMMA-SQL has been created as an improved extension of GEMMA-2B. For its improvements, GEMMA-SQL provide domain-specific fine-tuning on SQL datasets that greatly enhance accuracy in generating queries. Structural changes, including schema-aware encoding and SQL grammar constraints were also integrated to generate syntactically acceptable queries with appropriate context as well. GEMMA-SQL achieves a better balance of accuracy and computational efficiency, preserving the lightweight footprint of GEMMA-2B while outpacing larger proprietary models under conditions of limited resources. Finally, GEMMA-SQL employs schema linking methods to better generalize to unseen databases, and offers a collaborative and reproducible framework, unlike closed-source models. In tandem, these improvements present GEMMA-SQL as an accessible and reasonable text-to-SQL translation model. Table 4 presents a comparative overview of GEMMA-2B and GEMMA-SQL, highlighting their relative performance and key improvements.

**Table 4**. Comparative overview of GEMMA-2B and GEMMA-SQL

| Factors | Criteria | GEMMA-2B | GEMMA-SQL | Improvement in GEMMA-SQL |
|---|---|---|---|---|
| Accuracy | High accuracy on text-to-SQL tasks | ☐ | ☑ | Domain-specific fine-tuning on SQL datasets |
| Validity | Generates syntactically valid SQL queries | ☐ | ☑ | SQL grammar constraints + schema-aware encoding |
| Efficiency | Lightweight and suitable for low-resource devices | ☑ | ☑ | Retains 2B-scale efficiency while optimizing inference for SQL tasks |
| Generalization | Robust across unseen database schemas | ☐ | ☑ | Schema linking strategies for cross-database adaptability |

| Reproducibility | Open-source and reproducible for research | ☑ | ☑ | Benchmarked extensions remain open-source and replicable |
|---|---|---|---|---|
| Interpretability | Provides schema-aware reasoning transparency | ☐ | ☑ | Schema-awareness improves traceability of reasoning steps |
| Deployment | Ready-to-deploy for SQL applications | ☐ | ☑ | Pre-optimized pipeline for text-to-SQL with minimal customization |
| Scalability | Efficient scaling for domain-specific workloads | ☐ | ☑ | Task-focused scaling strategies to balance performance and resources |
| Resource Usage | Optimized resource-performance trade-off | ☐ | ☑ | Efficient fine-tuning reduces overhead without sacrificing accuracy |
| Openness | Open-source accessibility | ☑ | ☑ | Maintains open, transparent, and accessible framework |

Overall, both GEMMA-SQL and GEMMA-SQL Instruct demonstrate competitive and, in many cases, superior performance compared to a broad range of SOTA models. The results underscore the importance of instruction tuning for improving precision in SQL generation tasks. Despite having only 2 billion (2B) parameters, GEMMA-SQL consistently outperforms larger models, highlighting its efficiency and architectural sophistication. These findings position GEMMA-SQL Instruct as a highly effective and scalable solution for text-to-SQL applications.

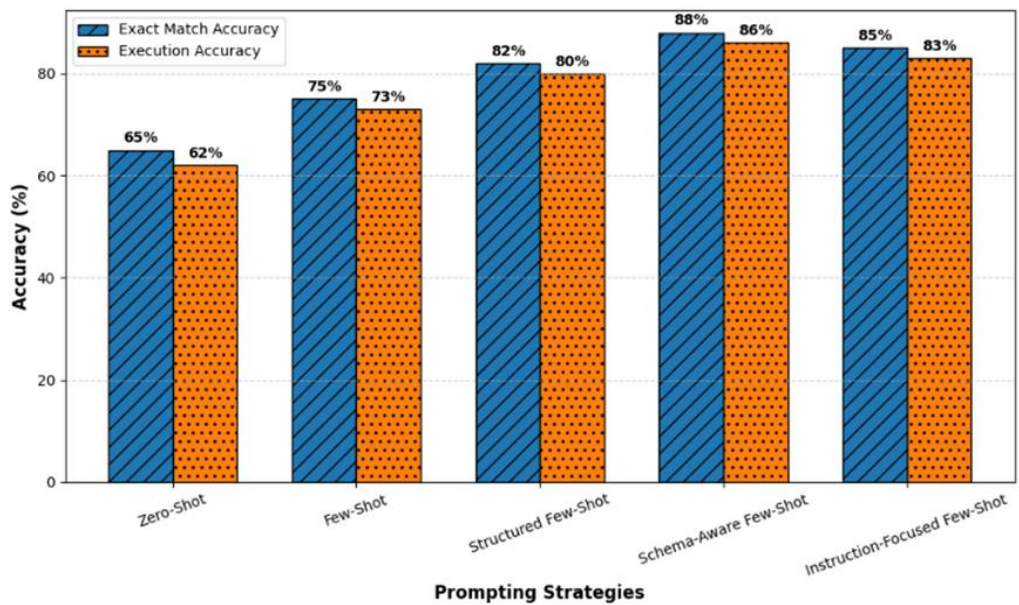

**Figure 22**. Prompt evaluation performance across different strategies.

### 4.4.2 Prompt Evaluation

We systematically evaluated GEMMA-SQL prompts on standard Text-to-SQL benchmarks using Exact Match (EM) and Execution Accuracy (EX) across both seen and unseen schemas. EM quantifies whether the generated SQL query exactly matches the reference query in syntax, including keywords, structure, and formatting, while EX measures whether the query produces identical results when executed on the database. Figure 22 compares EM and EX for five prompting strategies: zero-shot, few-shot, structured few-shot, schema-aware few-shot, and instruction-focused few-shot. The results show a consistent improvement with few-shot prompting. Zero-shot achieved the lowest accuracy (EM: 65%, EX: 62%), indicating limited capability without examples. Few-shot improved performance (EM: 75%, EX: 73%) by providing illustrative examples. Structured few-shot further increased accuracy (EM: 82%, EX: 80%) through an organized *instruction → schema → SQL* workflow. Schema-aware few-shot achieved the highest performance (EM: 88%, EX: 86%), highlighting the importance of explicit schema information. Instruction-focused few-shot maintained strong results (EM: 85%, EX: 83%) by emphasizing query intent and constraints. Overall, Figure 22 demonstrates that few-shot, structured, and schema-aware prompting strategies substantially enhance SQL generation accuracy and execution reliability compared to zero-shot approaches.

### 4.4.3 Analysis Summary

In summary, GEMMA-SQL and GEMMA-SQL Instruct demonstrate strong and consistent performance across a wide range of Text-to-SQL benchmarks, outperforming many state-of-the-art models despite having only 2 billion parameters. GEMMA-SQL achieves 64.5% TS accuracy and 60.7% EM, while GEMMA-SQL Instruct further improves these metrics to 66.8% TS and 63.3% EM, highlighting the effectiveness of instruction tuning. Comparative analyses across classical models, base LLMs, and instruction-tuned LLMs show that schema-aware and instruction-focused few-shot prompting strategies significantly enhance both syntactic and semantic SQL generation. Best-case scenarios demonstrate strong handling of complex nested queries, multi-table joins, and aggregation tasks, whereas worst-case examples reveal limitations in zero-shot performance and occasional semantic inconsistencies under unseen schemas. Overall, GEMMA-SQL balances efficiency, accuracy, and robustness, providing a practical, lightweight, and reproducible solution for cross-domain Text-to-SQL tasks, while the instruct variant further strengthens generalization and reliability through domain-aligned fine-tuning and structured prompting strategies.

## 5. Conclusions

In conclusion, this paper introduced GEMMA-SQL and GEMMA-SQL Instruct, two lightweight and efficient models for the Text-to-SQL task, developed using the open-source Gemma 2B architecture. Designed to translate natural language queries into executable SQL statements, these models address the dual challenges of performance and accessibility in semantic parsing. Empirical evaluations on the SPIDER benchmark highlight their effectiveness: GEMMA-SQL achieved 64.5% Test Suite (TS) accuracy and 60.7% Exact Match (EM), while GEMMA-SQL Instruct—its instruction-tuned variant—further improved performance to 66.8% TS and 63.3% EM. These results surpass several state-of-the-art baselines, including IRNet v2 (63.9% EM), RYANSQL (66.6% TS), CodeX-DaVinci (57.2% EM), and T5-base (57.2% EM), demonstrating the performance gains afforded by instruction tuning, even in compact models. A key advantage of the GEMMA-SQL family lies in its resource efficiency and open accessibility. Unlike many large-scale or closed-source counterparts, GEMMA-SQL models are open-source and optimized for deployment on low-resource hardware, making them highly suitable for academic, industrial, and edge applications. Their design—incorporating effective prompting mechanisms and lightweight post-processing—enables accurate and scalable SQL generation without the need for extensive computational infrastructure. Overall, GEMMA-SQL and GEMMA-SQL Instruct represent a promising direction for practical, high-performance semantic parsing. Their strong accuracy, low deployment overhead, and open accessibility make them ideal for real-world use cases where domain expertise or compute capacity may be limited. Future work will focus on domain adaptation, multilingual generalization, and integration with interactive querying systems to further expand the utility and impact of these models.

## 6. Acknowledgements

Authors would like to thank anonymous reviewers and our parent organizations for extending their support for the betterment of the manuscript.

## 7. Funding Information

This research did not receive any specific grant from funding agencies in the public, commercial, or not-for-profit sectors.

## 8. Conflict of Interest

The authors declare that they have no conflict of interest.

## 9. Data Availability Statement

The data supporting the findings of this study include both internal and external sources. The internal data (e.g., coding files) are available from the corresponding author, Hari Pandey, upon reasonable request. The primary external dataset used in this study is titled "Spider: A Large-Scale Human-Labeled Dataset for Complex and Cross-Domain Semantic Parsing and Text-to-SQL Task." This dataset is openly accessible through the "Spider 1.0 Yale Semantic Parsing and Text-to-SQL Challenge" repository. The other relevant materials can be found at: https://doi.org/10.5281/zenodo.16568737

## 10. Author Contributions

Anshul Gupta and Subham Sarkar contributed to the development and conceptualization of the research ideas, implemented the approach, performed experimental analysis, and participated in drafting and editing the manuscript. Minakshi Tomer contributed to idea development, provided guidance, and supported drafting and editing of the manuscript. Hari Mohan Pandey (corresponding author) led the conceptualization of the study, design the algorithms, supported to experimental and results analysis, provided overall direction and contributed to drafting and revising the manuscript. All authors have read and approved the final version of the manuscript. Johannes Schneider and Yan Gong supported in writing, reviewing and editing the manuscript.

## 11. ORCID

Anshul Gupta (0009-0006-9512-9318)
Subham Sarkar (0009-0002-6588-7389)
Minakshi Tomer (0000-0001-5536-3675)
Hari Mohan Pandey (0000-0002-9128-068X)
Schneider Johannes (0000-0002-6921-9049)
Yan Gong (0000-0003-2853-2108)